\documentclass{article}

\usepackage[final]{neurips_2024}

\usepackage[most]{tcolorbox}
\usepackage[utf8]{inputenc} 
\usepackage[T1]{fontenc}
\usepackage{url}            
\usepackage{booktabs}       
\usepackage{amsfonts}       
\usepackage{nicefrac}       
\usepackage{microtype}     
\usepackage{xcolor}       
\usepackage{graphicx}
\usepackage{adjustbox}
\usepackage{colortbl}
\usepackage{multirow}
\usepackage{paralist}
\usepackage{CJKutf8}
\usepackage{hyperref}
\usepackage{amsmath}
\usepackage{subcaption}
\usepackage{wrapfig}
\usepackage{todonotes}
\usepackage{enumitem}
\usepackage{setspace}

\definecolor{darkblue}{rgb}{0, 0, 0.5}
\hypersetup{colorlinks=true, citecolor=darkblue, linkcolor=darkblue, urlcolor=darkblue}
\definecolor{newblue}{RGB}{215,238,249}
\definecolor{neworange}{RGB}{255,243,221}
\definecolor{hidden-black}{RGB}{20,68,106}
\definecolor{tkcolor}{RGB}{255,243,221}
\newtcolorbox{takeaways}[2][]{
	width=\columnwidth,
	colback = tkcolor, 
	colframe = tkcolor, 
	boxsep=0pt,left=10pt,right=10pt,top=0pt,bottom=0pt,
	fontupper=\linespread{0.9}\selectfont,
	title=#2,#1}
\newtcolorbox{halftakeaways}[2][]{
	width=0.63\columnwidth,
	colback = tkcolor, 
	colframe = tkcolor, 
	boxsep=0pt,left=10pt,right=10pt,top=0pt,bottom=0pt,
	fontupper=\linespread{0.9}\selectfont,
	title=#2,#1}
\definecolor{shallowgray}{RGB}{237,240,246}
\definecolor{deepgray}{RGB}{117,117,117}
\newtcolorbox{prompt}[2][]{
	width=\columnwidth,
	colback = shallowgray, 
	colframe = black, 
	boxsep=2pt,left=10pt,right=10pt,top=5pt,bottom=5pt,
	fontupper=\linespread{1.5}\selectfont,
	title=#1}
\definecolor{newblue}{rgb}{0.85, 0.85, 0.9}
\title{What Factors Affect Multi-Modal In-Context Learning? An In-Depth Exploration}

\author{
	Libo Qin$^{\ddagger}$\thanks{Equal Contribution} \quad Qiguang Chen$^{\dagger}$\footnotemark[1] \quad Hao Fei$^\diamondsuit$ \quad Zhi Chen$^{\clubsuit}$ \quad Min Li$^{\ddagger}$ \quad Wanxiang Che$^{\dagger}$ \\
	$^{\ddagger}$ School of Computer Science and Engineering, Central South University \\
	$^{\dagger}$ Research Center for Social Computing and Information Retrieval\\
	$^\dagger$ Harbin Institute of Technology \quad $^\diamondsuit$ Tsinghua University \quad
	$^{\clubsuit}$ ByteDance \\
	\texttt{lbqin@csu.edu.cn}, \texttt{qgchen@ir.hit.edu.cn}
}

\begin{document}

	\maketitle

	\begin{abstract}
		Recently, rapid advancements in Multi-Modal In-Context Learning (MM-ICL) have achieved notable success, which is capable of achieving superior performance across various tasks without requiring additional parameter tuning. However, the underlying rules for the effectiveness of MM-ICL remain under-explored. To fill this gap, this work aims to investigate the research question: ``\textit{What factors affect the performance of MM-ICL?}''\ \ 
		To this end, we investigate extensive experiments on the three core steps of MM-ICL including demonstration retrieval, demonstration ordering, and prompt construction using 6 vision large language models and 20 strategies.
		Our findings highlight (1) the necessity of a multi-modal retriever for demonstration retrieval, (2) the importance of intra-demonstration ordering over inter-demonstration ordering, and (3) the enhancement of task comprehension through introductory instructions in prompts. We hope this study can serve as a foundational guide for optimizing MM-ICL strategies in future research.
	\end{abstract}
	
	\section{Introduction}
Recently, Large Language Models (LLMs) have demonstrated remarkable advancements, showcasing proficiency in a wide range of tasks~\citep{zhao2023survey,qin2023cross,qin2024large,hu2023tree,pan2023preliminary}.
Notably, advanced LLMs exhibit the emergence of novel capabilities such as In-Context Learning (ICL)~\citep{wei2022emergent,dong2022survey,zhuang2023through}, which optimize task performance by incorporating demonstrations into input prompts~\citep{pmlr-v202-looped,li2023transformers,wies2023learnability,zhou2022least}.
In particular, multi-modal in-context-learning (MM-ICL) is capable of utilizing multi-modal demonstrations to quickly adapt to the downstream task without parameter tuning~\citep{yin2023survey,he2023icl,zhang2024makes,li2024survey}.

In the literature, a series of works emerge  to enhance MM-ICL.
Specifically, \citet{gong2023multimodal} manually create a general template  with multiple images and corresponding responses during instruction-tuning (IT) stage to improve MM-ICL.
\citet{tsimpoukelli2021multimodal,li2023mimic,doveh2024towards} and \citet{zhao2024mmicl} develop task-specific MM-ICL templates during the IT stage, further extending its capabilities across more domains.
\citet{li2023otterhd} introduce OtterHD, adapting MM-ICL for high-definition image tasks.
Furthermore, \citet{sun2023generative} and \citet{tian2024mm} explore the potential of MM-ICL in the image generation tasks.
\citet{jin2024read} provide compelling evidence for the effectiveness of MM-ICL in comprehending game instructions. \citet{zong2024vl} and \citet{shukor2024beyond} develop fine-grained benchmarks and evaluate the MM-ICL in classification tasks.

While significant progress has been witnessed in MM-ICL, the existing work still mainly focuses on how to optimize MM-ICL, ignoring the underlying factors that influence its effectiveness and performance. Such gap impedes a comprehensive understanding of the mechanisms and performance determinants of MM-ICL, thereby limiting further exploration and research in this field. Motivated by this, this paper aims to  systematically investigate the research question: \textit{What factors affect the performance of MM-ICL?}, hoping to offer a unified view and guideline for researchers to build better MM-ICL. Specifically, as illustrated in Figure~\ref{fig:background}, the MM-ICL process comprises three steps: demonstration retrieval, demonstration ordering, and prompt construction. Therefore, We systematically investigate the following sub-questions: (a) \textit{how to select multi-modal demonstrations}  (Sec. \ref{sec:analysis-retrieval}); (b) \textit{how to order multi-modal demonstrations} (Sec. \ref{sec:analysis-order}); and (c) \textit{how to construct MM-ICL prompts} (Sec. \ref{sec:analysis-prompt}) to this end.
To achieve this, we conduct detailed experiments on MM-ICL using 20 strategies across 4 tasks with 6 representative vision large language models (VLLMs).

Through extensive investigations, the main findings are as follows:\vspace{-2mm}
\begin{itemize}[leftmargin=4ex]
	\item \textbf{Multi-modal alignment is the bottleneck for MM-ICL. }
	Our analysis confirms that, on average, multi-modal retrieval methods outperform single-modal ones. Furthermore, multi-modal alignment in VLLMs has a greater impact on MM-ICL effectiveness than parameter size, identifying alignment as the key limitation in both backbone structure and demonstration quality.
	\item \textbf{Intra-demonstration ordering holds greater importance than inter-demonstration ordering. } 
	Our investigation first indicates that the intra-demonstration ordering, particularly the ordering of modalities, greatly influences model performance more than demonstration arrangement.
	
	\item \textbf{Introductory instruction guides better task understanding for MM-ICL. }
	To construct a comprehensive MM-ICL prompt, it is essential to include introductory instructions preceding the demonstrations. This approach consistently enhances the performance of MM-ICL campared with summative instruction placed after demonstrations, and intra-demonstration instruction.\vspace{-3mm}
\end{itemize}

	\section{Background}\vspace{-2mm}
\begin{figure}[t]
	\centering
	\includegraphics[width=0.99\textwidth]{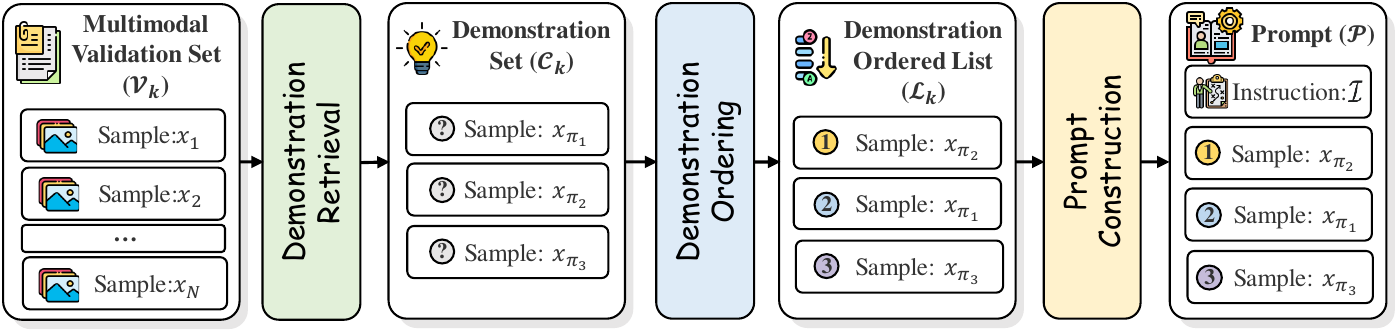}
	\caption{
			The whole process of prompting creation for multi-modal in-context-learning.\vspace{-3mm}
		}
	\label{fig:background}
\end{figure}
In this work, we formally present the prompt building process for MM-ICL.
As depicted in Figure~\ref{fig:background}, the process of prompt building for MM-ICL involves three sequential stages:

\textbf{(1) Demonstration Retrieval:} The core MM-ICL requires retrieval to obtain demonstrations that can help MM-ICL. Formally, given a validation dataset $\mathcal{V}_n = \{x_1, x_2, \dots, x_n\}$, each multi-modal sample $x_i$ includes textual input $I^{txt}_i$, visual input $I^{vis}_i$, and output $O_i$. For a specific test query $q$, this step aims to identify a subset of relevant demonstrations $\mathcal{C}_k = \{x_{\pi_j}\}_{j=1}^k$, where $x_{\pi_j} \in \mathcal{V}_n$.

\textbf{(2) Demonstration Ordering:}
Researches~\citep{lu-etal-2022-fantastically,wu2023self,xiang2024addressing} show that LLMs are highly sensitive to the order of demonstrations. Thus, arranging these demonstrations effectively is crucial for MM-ICL. After retrieving relevant demonstrations, we must rearrange the sequence $\mathcal{L}_k = [x_{\sigma_j}]_{j=1}^k$, which will be used to construct the prompt.

\textbf{(3) Prompt Construction:} Previous research indicates that using delimiters and instructions can significantly enhance textual ICL capabilities~\citep{min-etal-2022-rethinking,qin2023cross}. Therefore, the final core step is to transform the ordered demonstrations into a structured prompt $\mathcal{P}$, incorporating delimiters and instructions to optimize MM-ICL.\vspace{-2mm}
\section{What Factors Affect Multi-modal In-Context Learning?}\vspace{-1mm}
\label{sec:analysis}
\begin{figure}[t]
	\centering
	\includegraphics[width=0.99\textwidth]{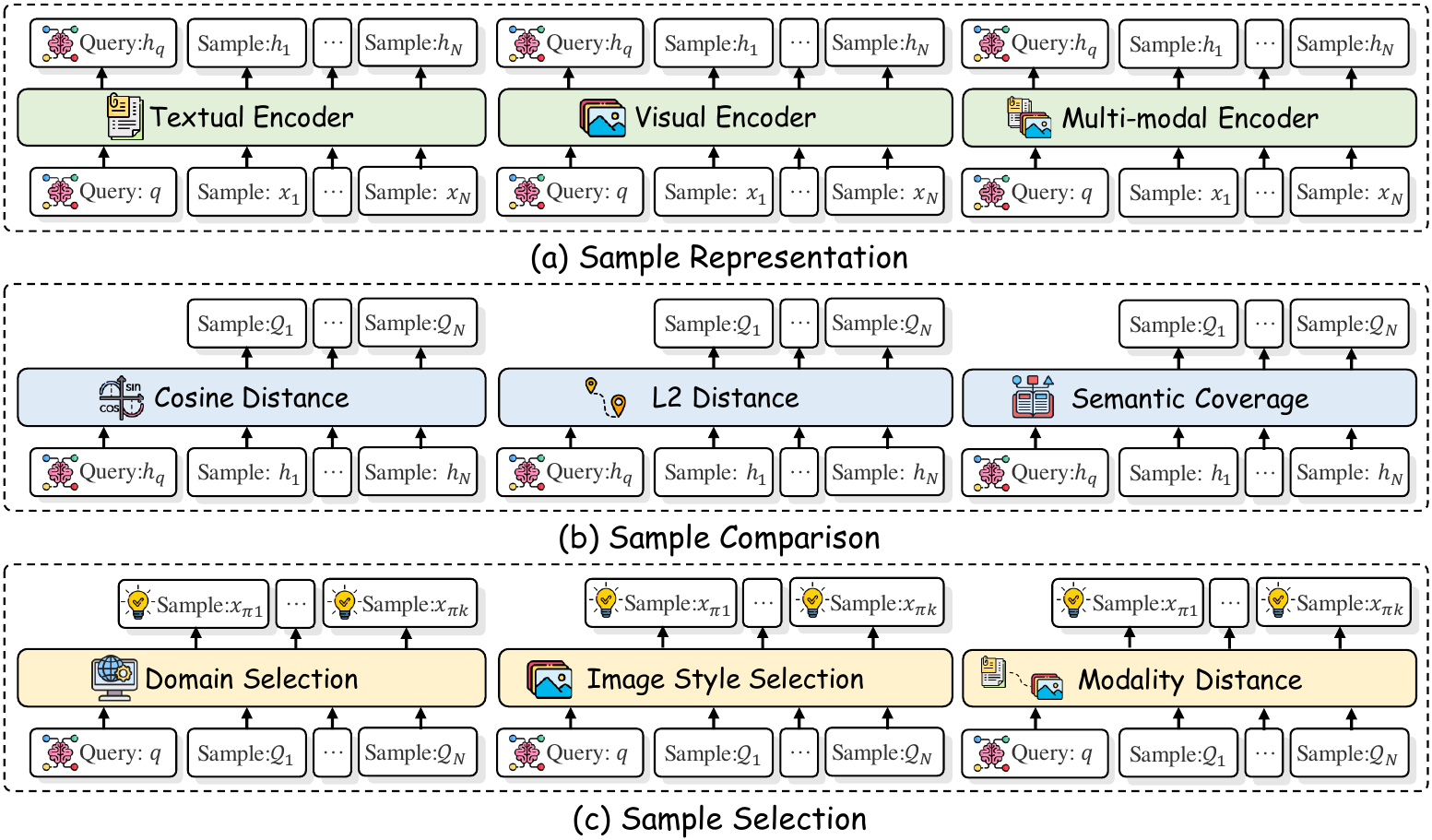}
	\caption{
		The demonstration retrieval process for MM-ICL.\vspace{-5mm}
	}
	\label{fig:demo-retrieval}
\end{figure}
\subsection{Exploration of MM-ICL Demonstration Retrieval}
\label{sec:analysis-retrieval}
The efficacy of ICL heavily depends on the quality of the retrieved demonstrations $\mathcal{C}$, which provide essential prior knowledge for MM-ICL. As illustrated in Figure~\ref{fig:demo-retrieval}, the retrieval process encompasses three key steps: {(1) Sample Representation, (2) Sample Comparison, and (3) Sample Selection}.
In this section, we conduct a systematic analysis of how various strategies for sample representation, comparison, and selection affect MM-ICL task performance.

\textbf{Sample Representation. } It involves defining an encoder ($\texttt{Encoder}(\cdot)$) to map each input sample $x_j \in \mathcal{V}$ and user query $q$ into a shared representation space:
\begin{equation}
	h_j = \texttt{Encoder}(x_j).
\end{equation}
Specifically, we evaluate various encoder architectures across modalities, focusing on the impact of visual encoder ($\texttt{Encoder}_{vis}$), text encoder ($\texttt{Encoder}_{txt}$), and multi-modal encoder ($\texttt{Encoder}_{multi}$) on model performance.

\textbf{Sample Comparison. } After deriving the representations, we employ a metric $\mathcal{M}$ to evaluate the quality $\mathcal{Q}_j$ of the sample $h_j$ in comparison to the query representation $h_q$ and the dataset samples $h_j$:
\begin{equation}
	\mathcal{Q}_j = \mathcal{M}(h_q, h_j).
\end{equation}
Specifically, we explore various comparison metrics, including cosine similarity $\mathcal{M}_{cos}$~\citep{liu-etal-2022-makes}, L2 similarity $\mathcal{M}_{L2}$~\citep{liu-etal-2022-makes}, and semantic diversity $\mathcal{M}_{div}$~\citep{li-qiu-2023-mot}, to assess sample quality and understand the correlation with model performance.

\textbf{Sample Selection. } After quality assessments, we apply a selection criterion $\mathcal{S}$ to identify the $k$ most advantageous samples $x_{\pi_j}$ for inclusion in the demonstration set $\mathcal{C}$:
\begin{equation}
	\mathcal{C} = \{x_{\pi_j}|x_{\pi_j} \in \mathcal{S}(q, \mathcal{Q}_j), j\le k\}.
\end{equation}
Sample selection is guided by factors such as domain information~\citep{he2023icl}, demonstration style~\citep{agrawal2023context}, and token distance~\citep{liu-etal-2022-makes}. Specifically, we systematically examine samples from both in-domain and out-of-domain collections. And we also assess the impact of image style on the selected demonstrations. Further, we investigate the token distance between modalities to understand its effects on sample selection for MM-ICL.

\subsection{Exploration of MM-ICL Demonstration Ordering}
\label{sec:analysis-order}
Following \citet{lu-etal-2022-fantastically} and \citet{wu2023self}, the order of the demonstration set $\mathcal{C}$ significantly impacts MM-ICL performance. As shown in Figure~\ref{fig:demo-ordering}, this section explores two key aspects:
\begin{figure}[h]
	\centering
	\includegraphics[width=0.99\textwidth]{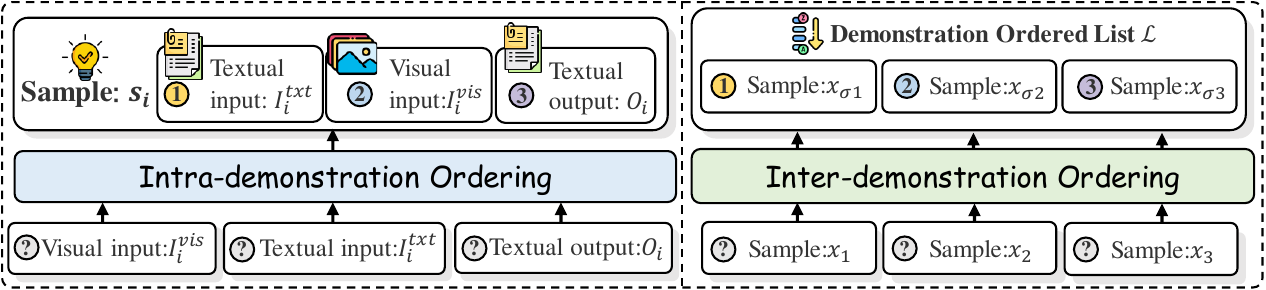}
	\caption{
		The demonstration ordering process for MM-ICL.\vspace{-5mm}
	}
	\label{fig:demo-ordering}
\end{figure}

\textbf{Intra-demonstration Ordering. } 
The sequence within a demonstration, especially modalities (e.g., text and image), is an important component that might affect the MM-ICL capabilities. Therefore, we introduce a intra-demonstration ordering permutation (\texttt{IOP}) to define this sequence:
\begin{equation}
	\mathcal{L} = [\texttt{IOP}(x_{\pi_1}),\texttt{IOP}(x_{\pi_2}),\ldots, \texttt{IOP}(x_{\pi_k})].
\end{equation}
We conduct a systematic exploration of various \texttt{IOP} configurations, including text-image-text ($\texttt{IOP}^{\texttt{tvt}}$), text-text-image ($\texttt{IOP}^{\texttt{ttv}}$), and image-text-text ($\texttt{IOP}^{\texttt{vtt}}$). These order analyses aim to evaluate the impact of different modal sequences on the model's performance.

\textbf{Inter-demonstration Ordering. } 
The sequence in which demonstrations are organized within $\mathcal{C}$ also is the key component that might impact the performance of MM-ICL. Formally, we define a sample ordering permutation $\sigma_j$ to specify the arrangement:
\begin{equation}
	\mathcal{L} = [x_{\sigma_1}, x_{\sigma_2}, \ldots, x_{\sigma_k}| x_{\sigma_j} \in \mathcal{C}],
\end{equation}
where $x_{\sigma_j}$ represents the $j$-th demonstration in the ordered demonstration list.

\subsection{Exploration of MM-ICL Prompt Construction}
\label{sec:analysis-prompt}

\begin{figure}[b]
	\centering
	\includegraphics[width=0.96\textwidth]{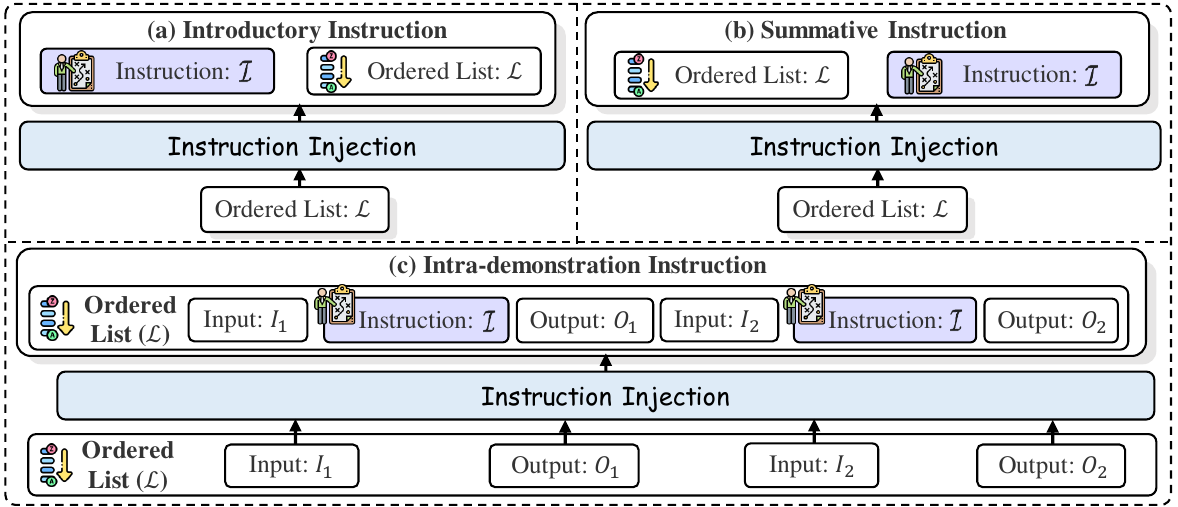}
	\caption{
		The process of instruction injection for MM-ICL prompt construction involves three key elements. The Introductory Instruction provides an overview instruction of the task before demonstrations. The Summative Instruction summarizes after the examples, guiding the model to apply the learned concepts to practical problems. The Intra-demonstration Instruction embeds task-specific guidance within each demonstration, enabling VLLMs to grasp task requirements during learning. Further details and additional prompts are provided in Appendix~\ref{append:instruction}.\vspace{-5mm}
	}
	\label{fig:prompt-construction-2}
\end{figure}
VLLMs are highly sensitive to input instructions~\citep{kojima2022large,qin2023cross}. Inspired by this, to enhance task comprehension, we incorporate different instructions to explore the performance influence for MM-ICL. Formally, we construct instruction methods $\mathcal{I}(\cdot)$ that describe the task and position them within the prompt. The prompt construction process is:
\begin{equation}
	\mathcal{P} = \mathcal{I}(\delta(x_{\sigma_1}), \delta(x_{\sigma_2}), \ldots, \delta(x_{\sigma_k})),
\end{equation}
Specifically, as shown in Figure~\ref{fig:prompt-construction-2}, we explore three instruction categories to bolster MM-ICL process:
\begin{itemize}[leftmargin=4ex]
	\item \textbf{Introductory Instruction ($\mathcal{I}_{intro}$)} refers to the initial guidance that offers an overview of the task prior to any demonstrations. As shown in Figure~\ref{fig:prompt-construction-2} (a), this instruction, denoted as $\mathcal{I}_{intro}$, is positioned at the start of the ordered demonstration sequence, $\mathcal{L}$.
	\item \textbf{Summative Instruction ($\mathcal{I}_{sum}$)} offers a summary after the examples, guiding the model to apply the learned concepts to real-world problems. As shown in Figure~\ref{fig:prompt-construction-2} (b), this instruction $\mathcal{I}$ is added at the end of the demonstration list $\mathcal{L}$.
	\item \textbf{Intra-demonstration Instruction ($\mathcal{I}_{intra}$)} embeds task instructions within each demonstration, helping VLLMs understand the task requirements during the learning process. As shown in Figure~\ref{fig:prompt-construction-2} (c), this instruction $\mathcal{I}$ is included within each demonstration $x_i$ in the list $\mathcal{L}$.
\end{itemize}
	\section{Experimental Setup}
\label{sec:setting}
Following the setting of \citet{li2023m}, we systematically explore 4 tasks, including image-caption, visual question answering (VQA), image classification, and chain-of-thought reasoning, which come from M$^3$IT~\citep{li2023m} and M$^3$CoT~\citep{chen-etal-2024-m3cot} (as shown in Tables~\ref{append:dataset}), providing a universal paradigm can help researchers conduct unified and fairer comparisons and studies within a unified framework.
In order to evaluate the MM-ICL performance accurately, we use two indicators for each task. Following \citet{zhang2019bertscore,li2023mimic}, and \citet{zong2024vl}, we use CIDER~\citep{vedantam2015cider} and BertScore~\citep{zhang2019bertscore} as image-caption metrics. Since M$^3$IT includes various VQA tasks with free-form answers, inspired by the success of free-form and precise answer hybrid evaluation in machine reading comprehension, following \citet{rajpurkar2016squad,zhang2019bertscore}, we adapt Token-F1~\citep{rajpurkar2016squad} and BertScore as visual question answering (VQA) metrics (The correlation analysis of the indicators and accuracy as shown in Table~\ref{append:correlation}). Following \citet{li2023m,li2023mimic}, we use accuracy and F1 score as indicators of image classification. Following \citet{lu2022learn,golovneva2022roscoe} and \citet{qin2023cross}, we use accuracy and reasoning alignment score~\citep{golovneva2022roscoe} (RAS) as  indicators of reasoning.

To ensure rigorous experimental control, we established a baseline using a multi-modal encoder for data representation and cosine similarity for sample comparison, limiting retrieval to within the same task.
This baseline ranks samples based on similarity, with a delimiter and a 3-shot setting (see Appendix~\ref{append:baseline} for details). 
In addition, all open source models complete inference on 2 A100 80G.
For all experiments, we select top-p from $\{0.95, 1\}$ and adjust the temperature parameter within $[0,1]$. Among them, temperature is the main error variable in this work.

\section{Empirical Analysis of Factors Affecting MM-ICL}
\subsection{Empirical Analysis of MM-ICL Demonstration Retrieval}
\begin{table*}[t]
	\centering
	\begin{adjustbox}{width=0.98\textwidth}
		\begin{tabular}{lccccccccc}
			\toprule
			\multirow{2}{*}{}  & \multicolumn{2}{c}{Caption} & \multicolumn{2}{c}{VQA} & \multicolumn{2}{c}{Classification} & \multicolumn{2}{c}{Reasoning} & \multirow{2}{*}{AVG}
			\\\cmidrule{2-9}
			& CIDER & BERTScore & Token F1 & BERTScore &  Acc & F1 & Acc &  RAS &
			\\
			\midrule
			\rowcolor{gray!8}\multicolumn{10}{c}{\textit{OpenFlamingo (9B)}~\citep{awadalla2023openflamingo}}\\
			\midrule
			\texttt{Zero-shot} & 1.84 & 81.18 & 2.78 & 76.17 & 15.17 & 3.63 & 16.53 & 85.13 & 35.30 
			\\
			\texttt{Few-shot (Random)} & 8.23 & 56.63 & 12.63 & 67.37 & 13.11 & 5.11 & 21.35 & 86.53 & 33.87 
			\\
			\texttt{ + Textual Retriever} & 13.39 & 74.22 & \textbf{21.80} & 75.74 & 13.67 & \textbf{12.04} & \textbf{25.63} & 87.71 & 40.52 
			\\
			\texttt{ + Visual Retriever} & 6.33 & 53.88 & 12.82 & 68.76 & 13.67 & 10.87 & 23.10 & 87.36 & 34.60 
			\\
			\texttt{ + Multi-Modal Retriever} & \textbf{13.47} & \textbf{85.01} & 7.85 & \textbf{79.05} & \textbf{19.66} & 10.10 & 24.96 & \textbf{88.11} & \textbf{41.03}
			\\
			\midrule
			\rowcolor{gray!8}\multicolumn{10}{c}{\textit{Otter (9B)}~\citep{li2023mimic}}\\
			\midrule
			\texttt{Zero-shot} & 2.86 & 86.42 & 20.90 & 87.95 & 24.34 & 10.85 & 34.06 & 82.67 & 43.76 
			\\
			\texttt{Few-shot (Random)} & 3.50 & 86.62 & 20.95 & 87.76 & 25.66 & 10.28 & 34.23 & 83.67 & 44.08 
			\\
			\texttt{ + Textual Retriever} & 3.89 & 86.62 & 20.40 & 87.89 & 25.28 & 8.47 & 32.88 & 81.93 & 43.42 
			\\
			\texttt{ + Visual Retriever} & 3.50 & 86.51 & 18.57 & 87.58 & 26.78 & 12.44 & 32.21 & 84.17 & 43.97 
			\\
			\texttt{ + Multi-Modal Retriever} & 3.77 & 86.57 & 18.80 & 87.56 & 28.65 & 11.83 & 35.92 & 83.74 & 44.60 
			\\
			\midrule
			\rowcolor{gray!8}\multicolumn{10}{c}{\textit{Qwen-VL (10B)}~\citep{Qwen-VL}}\\
			\midrule
			\texttt{Zero-shot} & 13.57 & 87.65 & 24.96 & 85.09 & 50.19 & 54.28 & \textbf{48.40} & 90.87 & 56.87 
			\\
			\texttt{Few-shot (Random)} & 28.52 & 88.47 & 28.43 & 86.11 & 52.43 & 53.50 & 43.34 & 90.19 & 58.87 
			\\
			\texttt{ + Textual Retriever} & 21.58 & 88.07 & 26.99 & 85.62 & 49.44 & 53.04 & 46.04 & 90.72 & 57.69 
			\\
			\texttt{ + Visual Retriever} & 30.81 & 88.56 & 28.79 & 86.23 & 59.74 & \textbf{54.43} & 46.88 & 91.30 & 60.84 
			\\
			\texttt{ + Multi-Modal Retriever} & \textbf{41.51} & \textbf{89.03} & \textbf{30.20} & \textbf{86.78} & \textbf{59.36} & 53.17 & 46.21 & \textbf{91.49} & \textbf{62.22} 
			\\
			\midrule
			\rowcolor{gray!8}\multicolumn{10}{c}{\textit{GPT4V (>100B)}~\citep{openai2023gpt4}}\\
			\midrule
			\texttt{Zero-shot} & 5.15 & 85.43 & 20.01 & 84.77 & 61.42 & 59.07 & 54.64 & 92.46 & 57.87 \\
			\texttt{Few-shot (Random)} & 6.37 & 85.95 & 24.43 & 85.42 & 60.11 & 60.81 & 54.30 & 92.54 & 58.74 
			\\
			\texttt{ + Textual Retriever} & 9.48 & 86.02 & 31.81 & \textbf{87.02} & 62.55 & 51.40 & 55.99 & 92.26 & 59.57 
			
			\\
			\texttt{ + Visual Retriever} & 9.36 & 86.26 & 32.47 & 86.96 & \textbf{63.30} & 57.79 & 59.87 & \textbf{93.19} & 61.15 
			
			\\
			\texttt{ + Multi-Modal Retriever} & \textbf{16.55} & \textbf{86.77} & \textbf{32.92} & 86.87 & 62.55 & \textbf{59.97} & \textbf{60.88} & 93.10 & \textbf{62.45}
			
			\\
			\midrule
			\rowcolor{gray!8}\multicolumn{10}{c}{\textit{IDEFICS2 (8B)}~\citep{laurenccon2024matters}}\\
			\midrule
			\texttt{Zero-shot} & 32.80 & 88.59 & 26.88 & 86.99 & \textbf{66.85} & 57.84 & \textbf{54.97} & 89.01 & 62.99 
			\\
			\texttt{Few-shot (Random)} & 39.68 & 88.88 & 30.82 & 87.59 & 61.61 & 53.39 & 51.94 & 89.52 & 62.93 
			\\
			\texttt{ + Textual Retriever} & 35.95 & 88.45 & 31.66 & 87.58 & 61.99 & \textbf{67.13} & 45.03 & 88.98 & 63.34 
			\\
			\texttt{ + Visual Retriever} & 46.61 & 89.55 & 32.05 & 87.92 & 64.04 & 62.51 & 51.26 & \textbf{89.83} & 65.47 
			\\
			\texttt{ + Multi-Modal Retriever} & \textbf{52.55} & \textbf{89.66} & \textbf{33.65} & \textbf{88.17} & 65.54 & 63.86 & 51.43 & 89.57 & \textbf{66.80}
			\\
			\midrule
			\rowcolor{gray!8}\multicolumn{10}{c}{\textit{Gemini-Pro (>100B)}~\citep{google2023gemini}}\\
			\midrule
			\texttt{Zero-shot} & 14.05 & 87.07 & 26.93 & 85.78 & 68.20 & 66.10 & 55.14 & 90.72 & 61.75 
			
			\\
			\texttt{Few-shot (Random)} & 21.21 & 88.13 & 32.99 & 86.81 & 63.67 & 69.75 & 55.65 & 91.82 & 63.75 
			\\
			\texttt{ + Textual Retriever} & 15.79 & 87.75 & 34.96 & 87.18 & \textbf{69.10} & \textbf{72.31} & 53.29 & 91.57 & 63.99 
			\\
			\texttt{ + Visual Retriever} & 21.35 & 87.98 & 44.74 & 89.33 & 65.73 & 64.18 & 53.29 & 91.92 & 64.81
			\\
			\texttt{ + Multi-Modal Retriever} & \textbf{35.64} & \textbf{88.67} & \textbf{45.47} & \textbf{89.61} & 65.17 & 70.51 & \textbf{58.01} & \textbf{92.17} & \textbf{68.16}  \\
			\bottomrule
		\end{tabular}
	\end{adjustbox}
	\caption{
		Performance comparison of retrievers utilizing different modal representations, where \texttt{Few-shot (Random)} refers to MM-ICL methods in which the demonstrations are randomly selected from the development set.\vspace{-8mm}
	}
	\label{exp:main-exp}
	
\end{table*}
\subsubsection{Sample Representation}
\textbf{Multi-modal alignment is the bottleneck for MM-ICL in both backbones and demonstrations.}
To evaluate the impact of semantic representation in different modalities for MM-ICL, we assessed three distinct encoders: RoBERTa~\citep{liu2019roberta} as a textual encoder for \texttt{Textual Retriever}, CLIP-Vision Encoder~\citep{radford2021learning} for \texttt{Visual Retriever}, and BridgeTower~\citep{xu2023bridgetower} as multi-modal encoder for  \texttt{Multi-Modal Retriever}.
As illustrated in Table~\ref{exp:main-exp}, multi-modal retrieval consistently outperforms zero-shot, randomly selected, and single-modality methods, highlighting the advantages of multi-modal semantic learning for MM-ICL. What's more, as shown in Table~\ref{exp:main-exp}, our results reveal that increasing model parameters from 8 billion to over 100 billion does not significantly enhance performance,
suggesting that beyond parameter size, multi-modal context understanding and alignment are more crucial for MM-ICL than model scaling. Our analysis demonstrates that multi-modal alignment is the critical factor in both the backbone and demonstrations.

\textbf{Current multi-modal encoders still lack modeling of multi-modal logic.} Actually, multi-modal retrieval attains better performance in many scenarios like Image Caption and VQA. However, our experiments show that textual retrieval works well for classification and reasoning tasks.
Based on the qualitative analysis, we observe that due to the semantic richness of the labels and rationales, textual retrieval can obtain more similar samples. However, the current multi-modal retrieval struggles with complex text semantics, often favoring image similarity. This aligns with recent work~\citep{tong2023massproducing,tong2024eyes,fei2024enhancing}, which is valuable for future exploration.

\textbf{Multi-modal context diminishes the necessity of careful demonstration selection.}
As shown in Table~\ref{exp:main-exp}, adding relevant demonstrations slightly improves performance, but the gains are less significant compared to text-only ICL scenarios. Specifically, retrieved demonstrations yield an average performance boost of 3.84\%, compared to random demonstrations. In contrast, text-only scenarios show performance increases of over 10\% with carefully selected samples~\citep{shi2023large}. Furthermore, the model remains unaffected by irrelevant samples, and the performance of almost all models is higher than zero-shot. This indicates that multi-modal context significantly reduces the need for careful demonstration selection, unlike in text-only scenarios.

\textbf{VLLMs learn semantic representations instead of token pattern representations for MM-ICL.}
As depicted by \citet{agrawal2023context}, textual ICL primarily learns token patterns (e.g., similar output formats, reasoning paths) among demonstration outputs. To investigate whether VLLMs rely on repetitive token patterns, we utilize the average BLEU score across demonstration outputs as a representation of token repetition. Figure~\ref{fig:token} shows that only the image captioning task exhibits a positive correlation. In contrast, other tasks show a decline as BLEU scores exceed 30\%. This underscores that MM-ICL primarily learns semantic rather than token pattern representations for effective performance.

\subsubsection{Sample Comparison}
To further analyze the influencing factors of MM-ICL in sample retrieval, this study employs similarity and diversity metrics, which help assess how MM-ICL processes sample similarities and differences, enhancing our understanding of its mechanisms.  See Appendix~\ref{append:sample-comarison} for more details and results.

\begin{figure}[t]
	\centering
	\begin{minipage}{0.38\textwidth}
		\centering
		\includegraphics[width=0.99\linewidth]{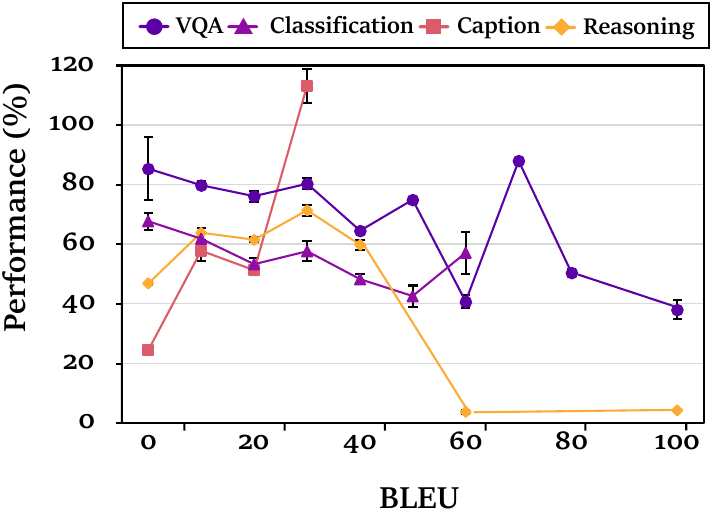}
		
		\captionof{figure}{The impact of token pattern representation in Gemini-Pro.
		}
		\label{fig:token}
	\end{minipage}
	\hfill
	\begin{minipage}{0.58\textwidth}
		\centering
		\includegraphics[width=0.99\linewidth]{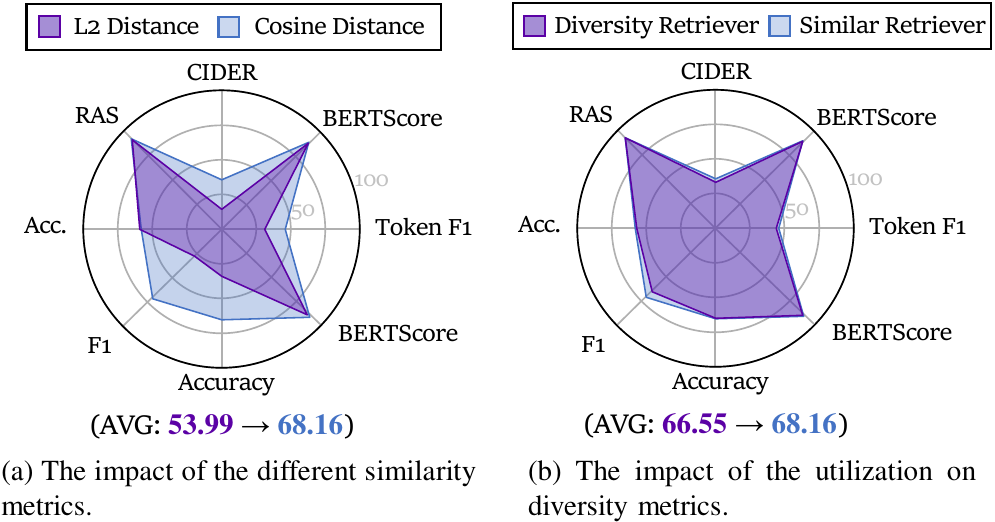}
		
		\captionof{figure}{The impact of different sample comparison methodologies in Gemini-Pro.}
		\label{fig:retriever}
	\end{minipage}
	\vspace{-5mm}
\end{figure}

\textbf{Cosine similarity matters for sample comparison.}
Following \citet{liu2022makes}, we compare two representative similarity metrics, cosine similarity and L2 similarity. As shown in Figure~\ref{fig:retriever} (a), cosine similarity, which measures the directional semantic alignment, emerges as the superior metric in MM-ICL than L2 similarity. Supported by \citet{deza2009encyclopedia} and \citet{steck2024cosine}, it indicates that MM-ICL prioritizes semantic directional consistency over complete semantic alignment.

\textbf{Diversity does not show significant influence for sample comparison.}
\citet{he2023icl,li2023finding} have shown that demonstrations with better diversity can effectively improve textual ICL. To explore whether it exists in MM-ICL, following \citet{li2023finding}, we ultilize the ``diversity retriever'', which selects the top-10 samples and further chooses the best 3 samples based on semantic diversity to obtain a more diverse MM-ICL.
As demonstrated in Figure~\ref{fig:retriever} (b), although diversity significantly enhances performance in text-based ICL, our experiments show limited improvement in MM-ICL tasks. This suggests that diversity may not directly correlate with better MM-ICL.
\begin{figure}[b]
	\centering
	\includegraphics[width=0.95\textwidth]{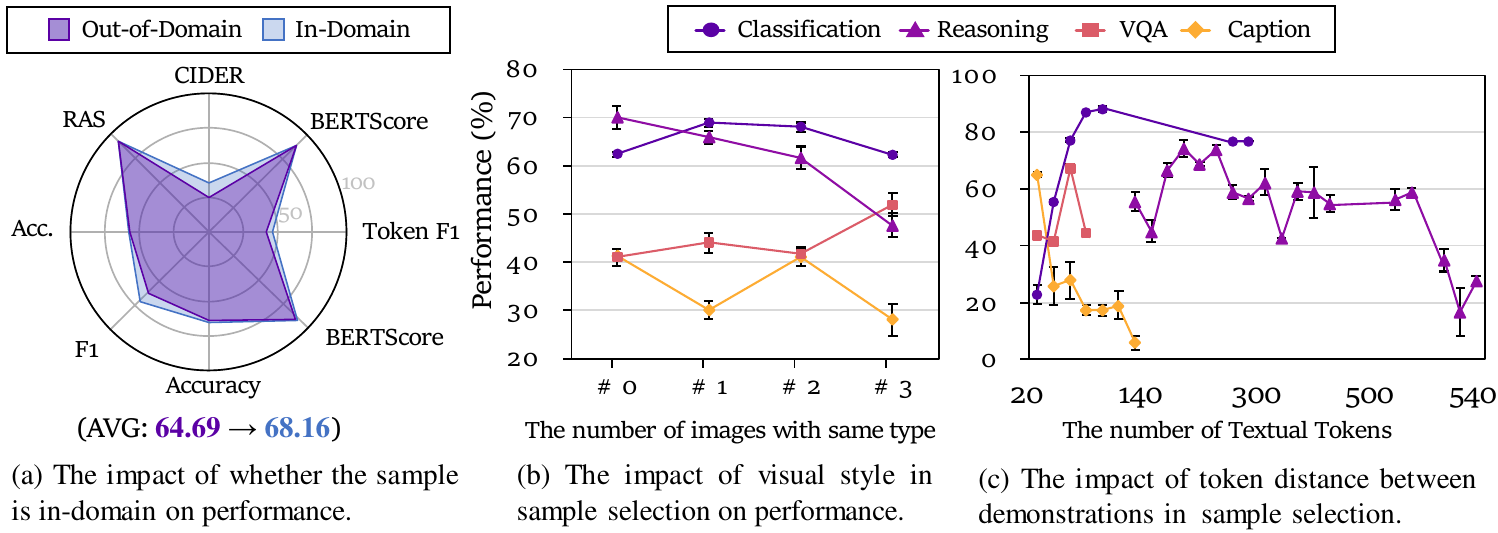}
	\caption{
		The impact of sample selection on average score performance in Gemini-Pro.
	}
	\label{fig:sample-selection}
\end{figure}

\subsubsection{Sample Selection}

\textbf{Domain interval matters for sample selection. }
Prior research highlights the critical role of domain relevance in enhancing ICL performance. Inspired by this, we employ the multi-modal retriever to select samples from both in-domain and out-of-domain pools. Figure~\ref{fig:sample-selection} (a) shows a nearly 4\% performance drop when out-of-domain demonstrations are included, underscoring the necessity of in-domain demonstrations for optimal MM-ICL.

\textbf{Visual style is not a crucial factor in sample selection. }
Although stylistic similarity in text samples is known to bolster ICL, its effect on the visual modality remains ambiguous. Utilizing CLIP for image classification, we investigate the impact of stylistic coherence in multi-modal samples on MM-ICL performance. As depicted in Figure~\ref{fig:sample-selection} (b), significant enhancements are observed solely in the VQA task, while captioning and classification show minimal effects and reasoning tasks decline. This indicates that diverse visual styles are not crucial in general MM-ICL.

\textbf{Token distances between modalities need to be considered for different tasks to improve sample selection. }
For textual ICL, excessive token distance between samples can impede performance~\citep{liu-etal-2022-makes}. We extend this inquiry to MM-ICL, analyzing how token distance across modalities influences results. Specifically, during the sample selection process, we considered the impact of the average token distance between two images on the model within the entire prompt of MM-ICL. As illustrated in Figure~\ref{fig:sample-selection} (c), the effect of token distance varies by task, typically showing an initial performance increase followed by a decline as distance grows, particularly in non-captioning tasks. This highlights the task-dependent nature of optimal token distance in MM-ICL.

\subsection{Empirical Analysis of MM-ICL Demonstration Ordering}
\begin{figure}[h]
	\centering
	\includegraphics[width=0.96\textwidth]{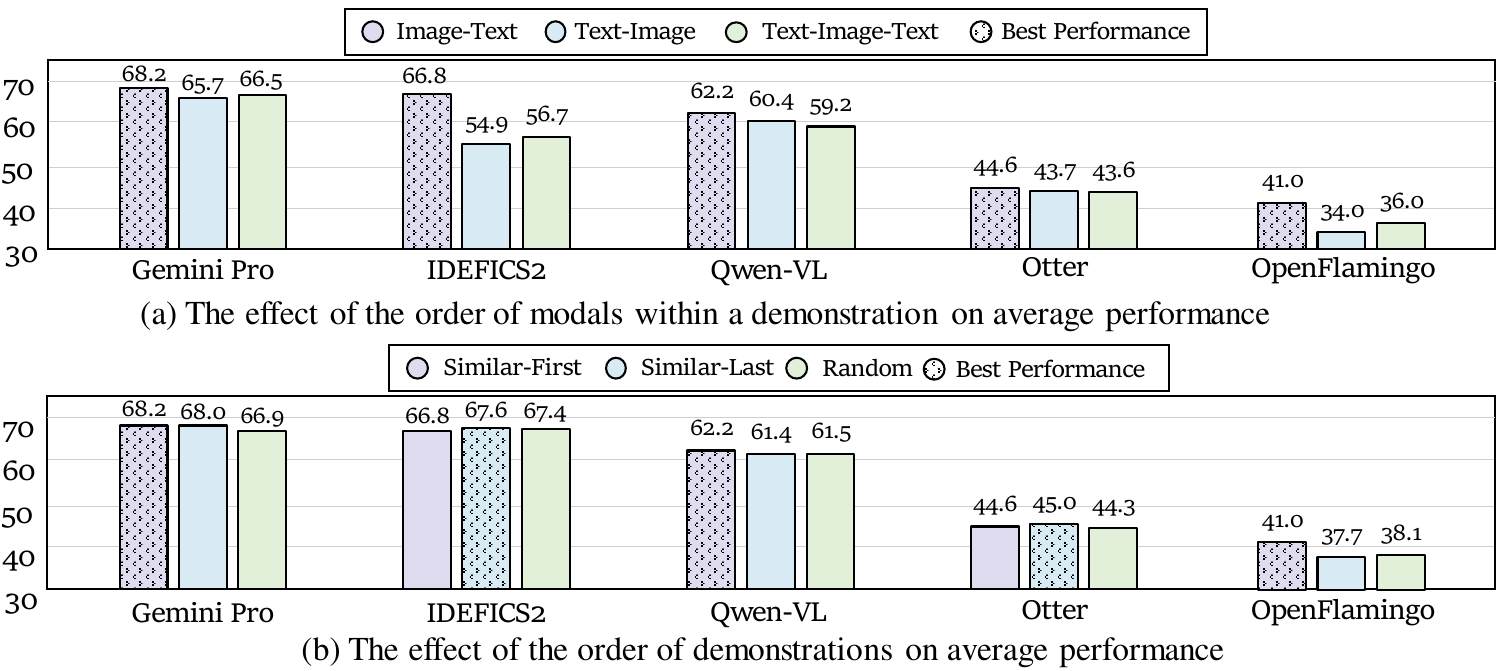}
	\caption{
		The impact of demonstration ordering on performance.
	}
	\label{fig:modal-order}
\end{figure}

\textbf{Intra-demonstration ordering significantly impacts performance.}
Within the demonstration, organizing the ordering, especially the relationship between modalities is a crucial topic. We investigate this by arranging inputs and outputs across modalities using three methods: \textit{text input$\rightarrow$text output$\rightarrow$image input} (Text-Image), \textit{text input$\rightarrow$image input$\rightarrow$text output} (Text-Image-Text), and \textit{image input$\rightarrow$text input$\rightarrow$text output} (Image-Text). As shown in Figure~\ref{fig:modal-order} (a), positioning the image at the start significantly enhances model performance. This suggests that presenting visual information first improves multi-modal comprehension, thereby boosting its learning abilities.

\textbf{Inter-demonstration ordering demonstrates minimal impacts.}
Following \citet{lu2022fantastically}, we investigate how the order of demonstration presentation influences model efficacy. We explore various strategies: random rearrangement, a "similar-last" approach where samples similar to the query are shown last, and a "similar-first" approach where similar samples are presented first. Figure~\ref{fig:modal-order} (b) illustrates that inter-demonstration ordering has a negligible impact on MM-ICL performance. This suggests the  order-robustness, with the presentation sequence having minimal effect.

\subsection{Empirical Analysis of MM-ICL Prompt Construction}
\begin{figure}[b]
	\centering
	\includegraphics[width=0.98\textwidth]{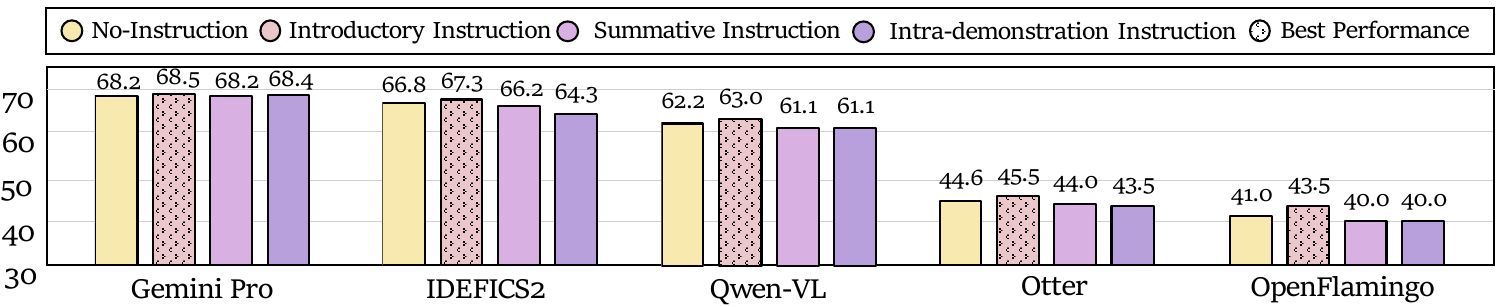}
	\caption{
		The impact of injecting instruction into demonstrations on model average score performance.
	}
	\label{fig:instruction}
\end{figure}
\begin{figure}[t]
	\centering
	\begin{minipage}{0.49\textwidth}
		\centering
		\includegraphics[width=0.99\linewidth]{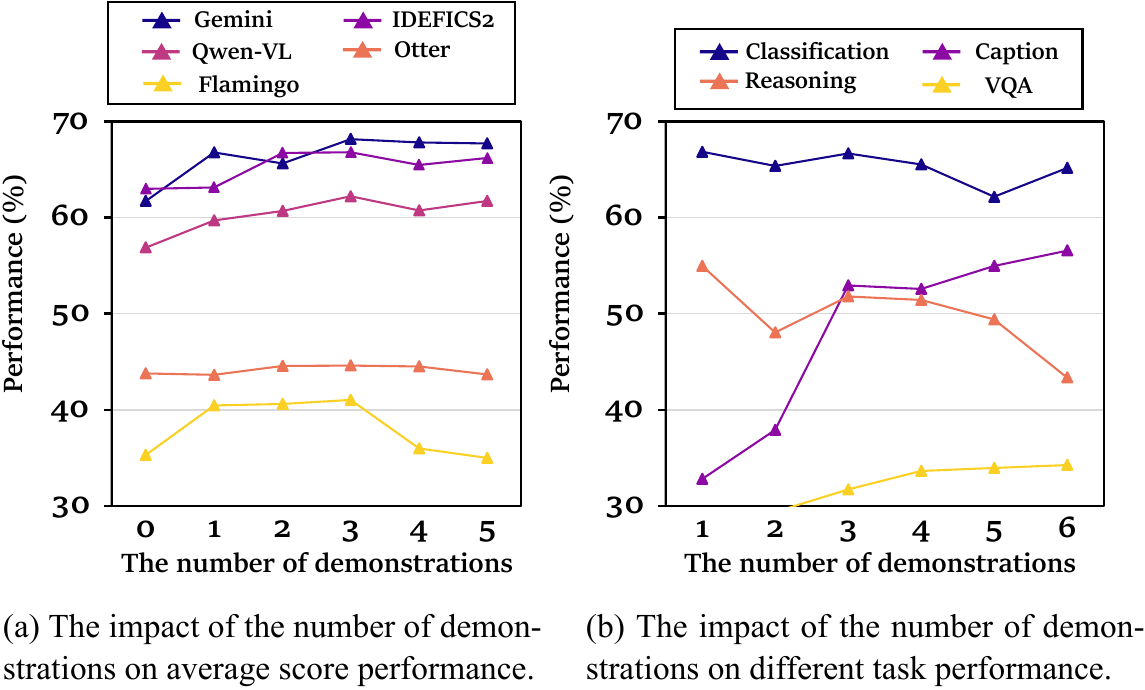}
		
		\captionof{figure}{
			The impact of the number of demonstrations on performance.
		}
		\label{fig:shot}
	\end{minipage}
	\hfill
	\begin{minipage}{0.49\textwidth}
		\centering
		\includegraphics[width=0.99\linewidth]{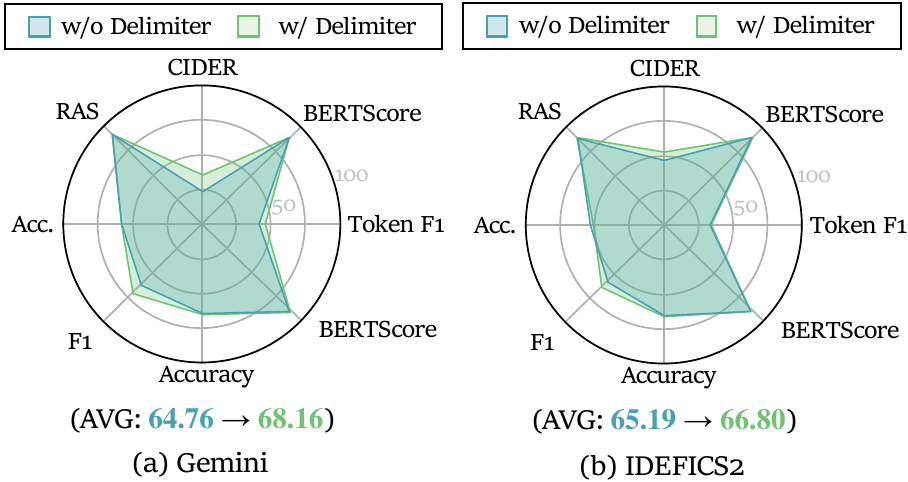}
		
		\captionof{figure}{
			The impact of inserting delimiter into the input and output of demonstration on model performance. See Figure~\ref{fig:tag-all} for more results.
		}
		\label{fig:tag}
	\end{minipage}
	\vspace{-3mm}
\end{figure}

\paragraph{Introductory Instruction is consistently effective for better MM-ICL.}
To investigate the impact of inserting task-related instructions within prompts, we conduct the following experiment on three categories of instruction:
\textit{Introductory Instruction},
\textit{Summative Instruction}, and
\textit{Intra-demonstration Instruction}.
As depicted in Figure~\ref{fig:instruction}, our analysis indicates that introductory instructions stably enhance model performance. In contrast, other instructions generally decrease performance. This finding suggests that introductory instructions facilitate targeted contextual learning and more effective semantic comprehension in demonstrations. We show more prompts and details in Appendix~\ref{append:instruction}.

\paragraph{MM-ICL is affected by the number of demonstrations depending on the task.}
Contrary to traditional text-based ICL, where performance improves with more samples, our findings in Figure~\ref{fig:shot} (a) suggest that MM-ICL does not experience significant gains from more demonstrations.
To further understand the reason behind, we analysis the performance on different tasks.
As shown in Figure~\ref{fig:shot} (b), increasing the number of demonstrations enhances performance in caption and VQA tasks, a trend also reported in prior studies~\citep{alayrac2022flamingo,laurenccon2024obelics,shukor2024beyond}. However, performance declines when demonstrations exceed three across all tested VLLMs. In more complex reasoning tasks, such as multi-step multi-modal chain-of-thought reasoning, additional demonstrations do not yield effective improvements, aligning with the findings of~\citet{chen-etal-2024-m3cot}, and \citet{fei2024video}.

Moreover, we attribute it to the following reasons for this limitation:
(1) \textbf{\textit{Cognitive Overload:}} For complex tasks, understanding numerous demonstrations can overwhelm the model, impeding its ability to process and integrate information effectively~\citep{chen2024unlocking}.
(1) \textbf{\textit{Complexity of Reasoning Tasks:}} In reasoning tasks, the performance improvement from more demonstrations is often less pronounced than when using diverse retrievers. This suggests that reasoning tasks require sophisticated integration of information, where quality outweighs quantity.
See Appendix~\ref{append:construction-sampling} for more detailed description.

\paragraph{The importance of delimiter lessens by text-image interleaved demonstrations.}
Previous research suggests that specific delimiters
for input and output data
can demonstrably influence textual ICL capabilities~\citep{min-etal-2022-rethinking}. 
Therefore, we utilize ablation experiments to omit these delimiters to examine their necessity (see Appendix~\ref{append:construction-delimiter} for details). As shown in Figure~\ref{fig:tag}, the resulting minor performance decline suggests that while these delimiters are less critical in MM-ICL, the modality switch inherent to MM-ICL may serve as an implicit delimiter, compensating for the absence of explicit delimiters.
	\section{Related Work}\vspace{-5pt}
\label{sec:visual-modal}
Recent advancements in vision large language models (VLLMs) have achieved great success in various vision-language tasks~\citep{yin2023survey,wu2024towards,wu24next,wang2024s3,fei2024vitron}. Initially, VLLMs lack Multi-modal In-context Learning (MM-ICL) capabilities.
To address this, researchers explore incorporating MM-ICL directly into the training phase. This involves constructing training samples with multi-modal interleaved data by manual and general templates, which unlock the MM-ICL capability~\citep{alayrac2022flamingo,awadalla2023openflamingo}. Building on this, \citet{li2023mimic,doveh2024towards} and \citet{zhao2024mmicl} extend the MM-ICL to construct  a series of task-specific templates, which improves generalization for MM-ICL. Further, \citet{li2023otterhd} introduce OtterHD and adapt the former process for high-definition images. The potential of MM-ICL is further explored in scene text recognition, image generation, and game instructions~\citep{zhao2023multi,sun2023generative,jin2024read}.

Recognizing the effectiveness of MM-ICL, researches shift towards prompt optimization. These methods focus on directly optimizing multi-modal prompts to understand the task and generate the expected output, without parameter adjustments~\citep{gong2023multimodal,tsimpoukelli2021multimodal,li2023mimic}. This approach has significantly improved performance in visual reasoning tasks~\citep{yang2022empirical,zheng2023ddcot}. Another approach involves textualizing visual information to enable VLLMs to leverage their background knowledge through in-context learning, further enhancing visual reasoning~\citep{yang2023mm, lu2024chameleon, gupta2023visual, shen2024hugginggpt}.
In addition, in order to better explore the MM-ICL, \citet{zong2024vl} and \citet{shukor2024beyond} also provide a dataset to test the MM-ICL capabilities of the multi-modal classification. Furthermore, \citet{shukor2024beyond} take the first step to conduct an instruction modification exploration for MM-ICL. 

Meanwhile, \citet{baldassini2024makes,chen2024understanding} pioneer the first naive multi-modal retrieval exploration to enhance MM-ICL.
Different from the existing work, our study mainly focuses on a \textbf{systematic exploration} of the effectiveness of key factors influencing the effectiveness of MM-ICL in a \textbf{unified perspective}.
To this end, we conduct a detailed analysis and exploration on 6 VLLMs and 20 factors across 4 tasks, aiming to provide systematic and practical guidance for future research.\vspace{-5pt}

	\section{Discussion}\vspace{-5pt}
\label{sec:discussion}
\paragraph{Broader Impacts.}
Our work is the first to systematically explore the factors influencing MM-ICL. We aim to enhance the understanding of MM-ICL mechanisms and guide future developments in this field. Additionally, our findings could foster a more comprehensive comprehension of MM-ICL within the community. For social impact, this research may influence the creation of more effective multi-modal large language models and relevant applications.

\paragraph{Limitations \& Future Work.} 
Due to time and cost constraints, this work is limited to the exploration of image and text modalities. In future research, we can extend our exploration to video modal ICL and multi-lingual MM-ICL scenarios. Another limitation of this work involves the insufficient consideration of certain image instructions, such as grounding or the inclusion of additional arrows. These aspects often require more complex human input and are not adequately supported by most current models.\vspace{-5pt}
\section{Conclusion}\vspace{-5pt}

This study is the first to systematically explore MM-ICL by identifying key performance determinants. Our experiments with 6 models and 20 factors across 4 tasks show that multi-modal retrieval significantly outperforms single-modal approaches and the intra-demonstration ordering critically influences learning efficacy. Additionally, incorporating task-specific instructions into prompts enhances model performance. We hope these findings will refine our understanding of MM-ICL mechanisms and guide more effective developments and future research in this evolving field.

\section*{Acknowledgments}
This work was supported by the National Natural Science Foundation of China (NSFC) via grant 62306342, 62236004, 62441603 and 62476073. This work was also sponsored by the Excellent Young Scientists Fund in Hunan Province (2024JJ4070) and the Science and Technology Innovation Program of Hunan Province under Grant 2024RC3024. We are grateful for resources from the High Performance Computing Center of Central South University, and the CCF-Zhipu.AI Large Model Innovation Fund. Libo Qin is the corresponding author.

\bibliographystyle{plainnat}
\bibliography{ref}

\newpage
\section*{Appendix}
\setcounter{subsection}{0}
\renewcommand{\thesubsection}{\Alph{subsection}}
\begin{table*}[b]
	\centering
	\begin{adjustbox}{width=0.6\textwidth}
		\begin{tabular}{l|c}
			\toprule
			Dataset & Category \\
			\midrule
			COCO Caption~\citep{chen2015microsoft} & IC \\
			TextCaps~\citep{sidorov2020textcaps} & IC \\
			Paragraph Captioning~\citep{krause2017hierarchical} & IC \\
			\midrule
			COCO Text~\citep{veit2016coco} & CLS \\
			ImageNet Image Classification~\citep{russakovsky2015imagenet} & CLS \\
			IQA~\citep{duanmu2021quantifying} & CLS \\
			COCO-ITM~\citep{chen2015microsoft} & CLS \\
			e-SNLI-VE~\citep{kayser2021vil} & CLS \\
			Mocheg~\citep{yao2023end} & CLS \\
			\midrule
			VQA-v2~\citep{goyal2017making} & VQA \\
			DocVQA~\citep{mathew2021docvqa} & VQA \\
			OCR-VQA~\citep{mishra2019ocr} & VQA \\
			ST-VQA~\citep{biten2019scene} & VQA \\
			Text-VQA~\citep{singh2019towards} & VQA \\
			GQA~\citep{hudson2019gqa} & VQA \\
			OKVQA~\citep{marino2019ok} & VQA \\
			A-OKVQA~\citep{schwenk2022okvqa} & VQA \\
			\midrule
			ScienceQA~\citep{lu2022learn} & R \\
			M$^3$CoT~\citep{chen-etal-2024-m3cot} & R \\
			\bottomrule
		\end{tabular}
	\end{adjustbox}
	\caption{
		Dataset in M$^3$IT and M$^3$CoT, where IC: Image Captioning, CLS: Classification, VQA: Visual Question Answering, R: Chain-of-Thought Reasoning (with NL rationale). Due to the cost, for each task, we evenly sampled 500 items according to the sub-dataset.
	}
	\label{append:dataset}
	
\end{table*}
\subsection{The Implement Details for Standard Baseline}
\label{append:baseline}

To ensure rigorous control of experimental variables, we establish a standard baseline for our study. This baseline utilizes a multi-modal encoder for data representation and cosine similarity for sample comparison, with retrieval restricted to the same task. The following sections provide detailed insights into the implementation of this baseline.
\subsubsection{Demonstration Retrieval Implementation for Baseline}
\paragraph{Multi-modal Encoder for Sample Representation}

We employ BridgeTower~\citep{xu2023bridgetower} as a multi-modal encoder to represent the data in a unified embedding space. This encoder integrates both visual and textual information, prioritizing single modality to capture the rich semantic content present in images.

\paragraph{Cosine Similarity for Sample Comparison}
To compare samples effectively, we use cosine similarity, a metric that measures the cosine of the angle between two non-zero vectors in a multi-dimensional space. This choice is motivated by its effectiveness in capturing the similarity between high-dimensional vectors, which are typical outputs of our multi-modal encoder. Specifically, we compute the cosine similarity between the query and each candidate sample, which is given by:
\[
\text{cosine}(h_q, h_i) = \frac{h_q \cdot h_i}{\|h_q\| \|h_i\|}
\]
where \( h_q \) and \( h_i \) are the embedding vectors of the query $q$ and candidate sample $x_i$, respectively.

\paragraph{In-domain and Top-k Retrieval for Sample Selection}
To ensure the relevance and accuracy of the retrieval process, for sample selection, we first confine retrieval to the same task and domain. This means that comparisons and rankings are conducted exclusively among samples within the same task and domain category, ensuring the contextual appropriateness of the retrieved results.

In addition, for sample selection, samples are ranked according to their cosine similarity scores. Higher similarity scores indicate a closer alignment with the query sample, enabling the efficient identification of the most relevant samples. This ranking process involves two main steps: (1) Sorting: Candidate samples are sorted in descending order based on their cosine similarity scores relative to the query. (2) Selection: Subsequently, the top-k ranked samples are selected based on their relevance as determined by the similarity scores.
\subsubsection{Demonstration Ordering Implementation for Baseline}
By default, we utilize the methodology for ordering demonstrations within our baseline model. By default, we adopt a text-after-image (\texttt{Text-Image}) approach for intra-demonstration sorting. This means that, within a single demonstration, textual information is positioned after the corresponding image. This ordering is chosen based on preliminary findings suggesting that such a sequence aids in better contextual understanding and retention of the demonstrated information.

Furthermore, for the ordering of inter-demonstration sequences, we employ a similarity-based method. This method ranks demonstrations according to their similarity to the query, with more similar demonstrations placed higher in the order. The similarity is determined using a metric that assesses the alignment of key features between the query and the demonstrations. This approach ensures that the most relevant and contextually aligned demonstrations are prioritized, potentially enhancing the model’s performance and the user’s comprehension.

\subsubsection{Prompt Construction Implementation for Baseline}
To ensure consistency and comparability in our baseline, we introduce both a delimiter and a 3-shot setting (following~\citet{wei2022chain,qin2023cross}). The delimiter serves to clearly demarcate different segments of the input data, preventing any potential confusion or overlap between distinct portions of the input. This clear separation is crucial for the model to accurately process and understand the structure of the data it receives.

The 3-shot setting, on the other hand, involves providing three examples for each task within the prompt. This approach is designed to stabilize the learning process by presenting the model with sufficient contextual information. By offering three examples, we strike a balance between providing enough context to guide the model's understanding and avoiding the cognitive overload that might occur with too many examples. This setting not only enhances the model's performance but also ensures a more robust and reliable learning process.

\subsubsection{Baseline Prompt}
In the context of using Vision-and-Language Large Models (VLLMs), it is essential to carefully structure the input prompts to ensure accurate processing. The prompt format typically used is illustrated below:
\begin{prompt}
	\ \ 
	\textbf{[REQUEST]} \textcolor{deepgray}{\% Shot 1}
	
	\ \ \textit{<Visual Input $\mathcal{I}_1^{vis}$>} \ \ \textit{<Textual Input $\mathcal{I}_1^{txt}$>}
	
	\textbf{[RESPONSE]}
	
	\ \ \textit{<Textual Output $\mathcal{I}_1^{vis}$>}
	
	\textbf{[REQUEST]} \textcolor{deepgray}{\% Shot 2}
	
	\ \ \textit{<Visual Input $\mathcal{I}_2^{vis}$>} \ \ \textit{<Textual Input $\mathcal{I}_2^{txt}$>}
	
	\textbf{[RESPONSE]}
	
	\ \ \textit{<Textual Output $\mathcal{I}_2^{vis}$>}
	
	\textbf{[REQUEST]} \textcolor{deepgray}{\% Shot 3}
	
	\ \ \textit{<Visual Input $\mathcal{I}_3^{vis}$>} \ \ \textit{<Textual Input $\mathcal{I}_3^{txt}$>}
	
	\textbf{[RESPONSE]}
	
	\ \ \textit{<Textual Output $\mathcal{I}_3^{vis}$> }
	
	\textbf{[REQUEST]} \textcolor{deepgray}{\% User Query}
	
	\ \ \textit{<Visual Input $\mathcal{I}_q^{vis}$>} \ \ \textit{<Textual Input $\mathcal{I}_q^{txt}$>}
	
\end{prompt}
where 
any gray text following the percent sign (\%) is treated as a comment. These comments are not processed as part of the primary input but serve to provide additional context or instructions within the coding environment. This convention helps in maintaining the clarity and functionality of the given prompting.

In conclusion, the standard baseline established here integrates a multi-modal encoder, cosine similarity, and task-specific retrieval with a focus on visual modalities. It ranks samples based on similarity and employs a delimiter with a 3-shot setting to ensure robust and consistent performance across different tasks.

\subsection{The Implement Details for Sample Comparison}
\label{append:sample-comarison}

\subsubsection{Metric Calculation}
\paragraph{Cosine Similarity ($\mathcal{M}_{cos}$)} Compute the cosine similarity between $h_q$ and $h_j$ using the formula:
\begin{equation}
	\mathcal{M}_{cos}(h_q, h_j) = \frac{h_q \cdot h_j}{\|h_q\| \|h_j\|}
\end{equation}

\paragraph{L2 Similarity ($\mathcal{M}_{L2}$)} Calculate the L2 similarity by computing the negative Euclidean distance between $h_q$ and $h_j$:
\begin{equation}
	\mathcal{M}_{L2}(h_q, h_j) = -\|h_q - h_j\|_2
\end{equation}
Since Euclidean distance measures dissimilarity, we use the negative value to represent similarity, where a higher value indicates greater similarity.

\paragraph{Semantic Diversity ($\mathcal{M}_{div}$)} Semantic diversity is assessed by evaluating the differences in the distributional properties of $h_q$ and $h_j$. This assessment involves analyzing the variance in how these properties are distributed across different samples. To determine the presence of semantic diversity within Multi-Modal In-Context Learning (MM-ICL), we adopt the methodology proposed by \citet{li2023finding}. Specifically, we employ the "diversity retriever," designed to enhance the diversity of the selected samples. The diversity retriever operates by first selecting the top 10 samples based on a preliminary measure of relevance. From these top 10 samples, it then identifies the 3 samples that exhibit the highest semantic diversity. This two-step process ensures that the final selection of samples for MM-ICL is not only relevant but also diverse in terms of their semantic content.

\subsubsection{Comparison and Analysis}

Comparing the results obtained using different metrics ($\mathcal{M}_{cos}$, $\mathcal{M}_{L2}$, $\mathcal{M}_{div}$) provides a comprehensive understanding of their effectiveness and suitability for specific applications. It is essential to analyze the trade-offs associated with each metric and interpret the results to draw meaningful conclusions about sample quality and relevance.

\begin{figure}[t]
	\centering
	\includegraphics[width=0.98\textwidth]{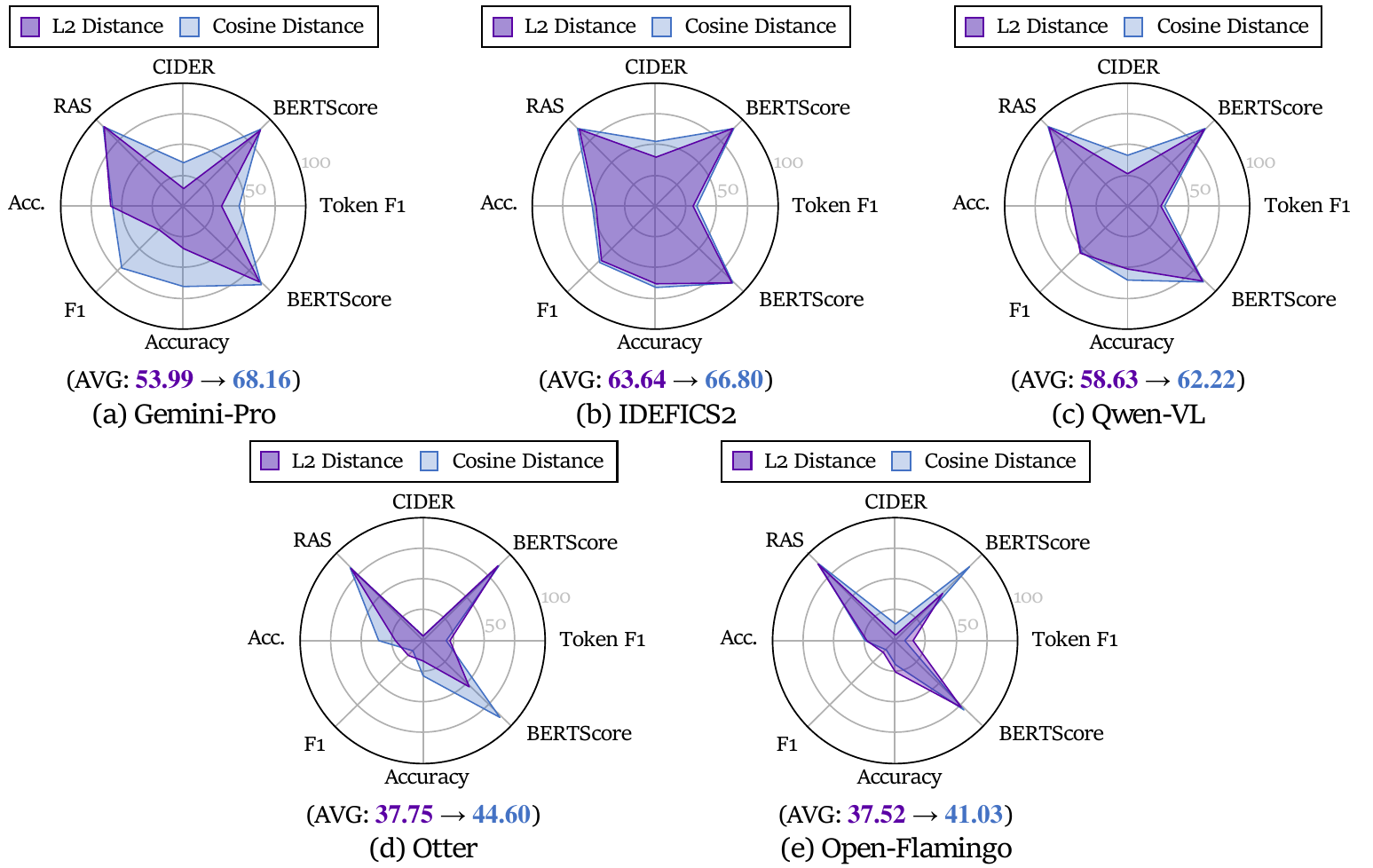}
	\caption{
		The impact of the different similarity metrics.
	}
	\label{fig:cos-retriever-all}
\end{figure}

As shown in Figure~\ref{fig:cos-retriever-all}, cosine similarity, which measures directional semantic alignment, emerges as the superior metric in MM-ICL compared to L2 similarity. This observation is supported by the findings of \citet{deza2009encyclopedia} and \citet{steck2024cosine}, who highlight that MM-ICL prioritizes semantic directional consistency over complete semantic alignment. Cosine similarity's ability to capture the nuances of directional alignment allows for more precise interpretations of semantic relationships within the data, making it particularly effective for MM-ICL tasks.

\begin{figure}[t]
	\centering
	\includegraphics[width=0.98\textwidth]{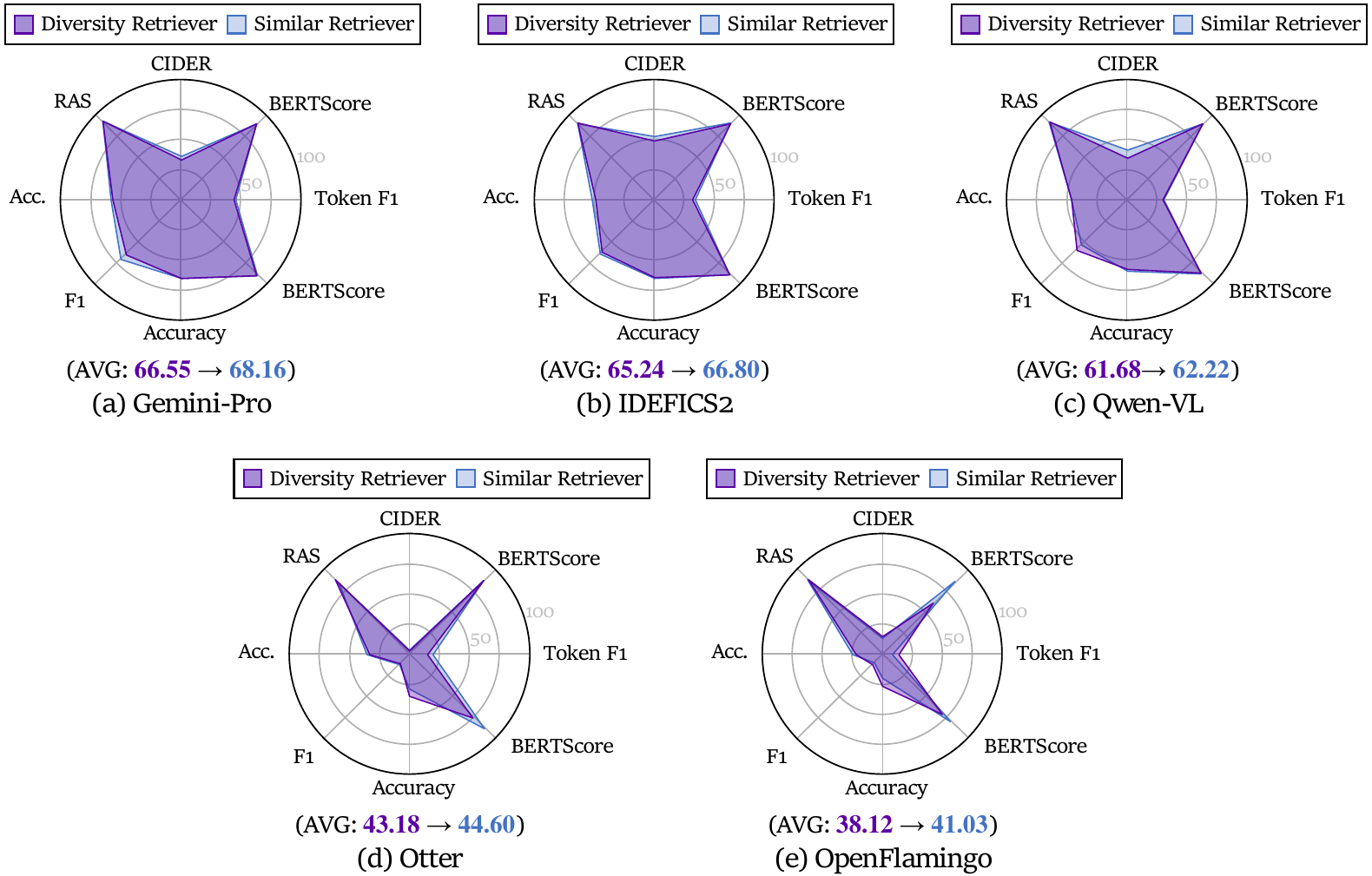}
	\caption{
		The impact of the utilization on diversity metrics.
	}
	\label{fig:diverse-retriever-all}
\end{figure}

In contrast, Figure~\ref{fig:diverse-retriever-all} illustrates that while diversity, as measured by $\mathcal{M}_{div}$, enhances performance in text-based in-context learning, our experiments reveal limited improvement in MM-ICL tasks. This finding suggests that diversity may not directly correlate with better performance in MM-ICL. The limited impact of diversity on MM-ICL performance could be attributed to the specific nature of multi-modal data, where the interplay between different modalities requires a more nuanced approach than simply maximizing diversity. 

Further analysis of these metrics reveals the inherent trade-offs between them. For instance, while cosine similarity offers advantages in maintaining semantic directional consistency, it may not capture the full extent of semantic similarity that L2 similarity can provide. On the other hand, L2 similarity, though comprehensive in measuring complete alignment, might lack the precision needed for tasks that rely heavily on directional semantic cues. Similarly, while diversity is beneficial in certain contexts, its role in MM-ICL needs to be reconsidered, potentially focusing on optimizing other aspects of sample quality.

In summary, the evaluation of $\mathcal{M}_{cos}$, $\mathcal{M}_{L2}$, and $\mathcal{M}_{div}$ underscores the importance of selecting appropriate metrics based on the specific requirements of the task. Understanding the trade-offs and context-specific effectiveness of these metrics is crucial for optimizing performance in multi-modal in-context learning applications.

\subsection{Exploration of MM-ICL Prompt Construction}
\label{append:construction}

\subsubsection{The Implement Details for Demonstration Sampling}
\label{append:construction-sampling}
\begin{figure}[h]
	\centering
	\includegraphics[width=0.99\textwidth]{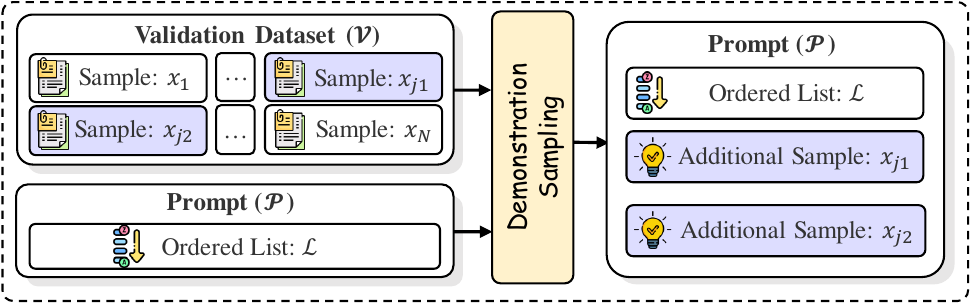}
	\caption{
		The demonstration sampling process for MM-ICL prompt construction.
	}
	\label{fig:prompt-construction-3}
\end{figure}

To examine the effect of demonstration sample quantity on model performance, as shown in Figure~\ref{fig:prompt-construction-3}, we select a subset of $k'$ demonstrations from the demonstration list $\mathcal{L}_{k'}$ to the prompt, where $k'$ is the number of retrieved demonstrations.
Formally, the prompt construction process is defined as:
\begin{equation}
	\mathcal{P} = \mathcal{I}(\delta(x_{\sigma^j_1}), \delta(x_{\sigma^j_2}), \ldots, \delta(x_{\sigma^j_{k'}}))
\end{equation}
We systematically evaluate the influence of varying $k'$ on MM-ICL performance.

\subsubsection{The Implement Details for Delimiter Injection}
\label{append:construction-delimiter}
\begin{figure}[h]
	\centering
	\includegraphics[width=0.99\textwidth]{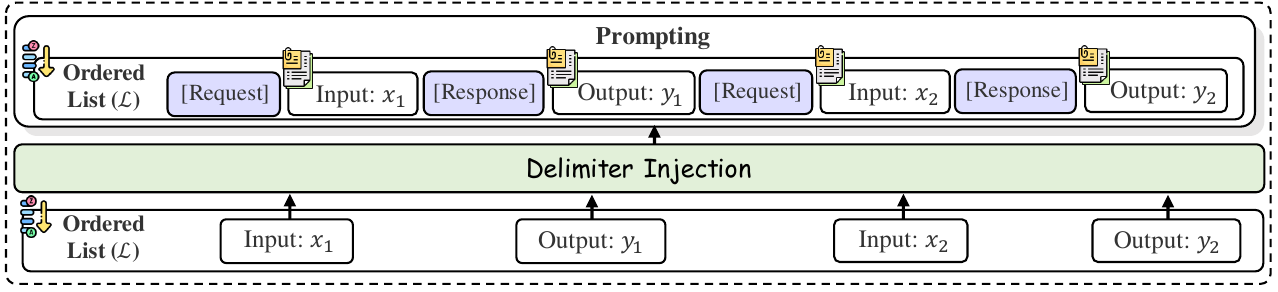}
	\caption{
		The delimiter injection process for MM-ICL prompt construction.
	}
	\label{fig:prompt-construction-1}
\end{figure}

To distinctly separate inputs and outputs within demonstrations $x_i$, as shown in Figure~\ref{fig:prompt-construction-1}, we leverage special delimiter  markers. Delimiters like \texttt{[Request]} and \texttt{[Response]} are strategically placed before the inputs and outputs, respectively. Formally, delimiter injection function $\delta$ maps inputs and outputs to the prompting sequences:
\begin{equation}
	\delta(x_{\sigma_i}) = \texttt{[Request]} \oplus I_i \oplus \texttt{[Response]} \oplus O_i,
\end{equation}
where $I_i$ and $O_i$  denotes the input and output for the sample $x_i$, respectively. In addition, $\oplus$ represents string concatenation operation.
\begin{figure}[h]
	\centering
	\includegraphics[width=0.98\textwidth]{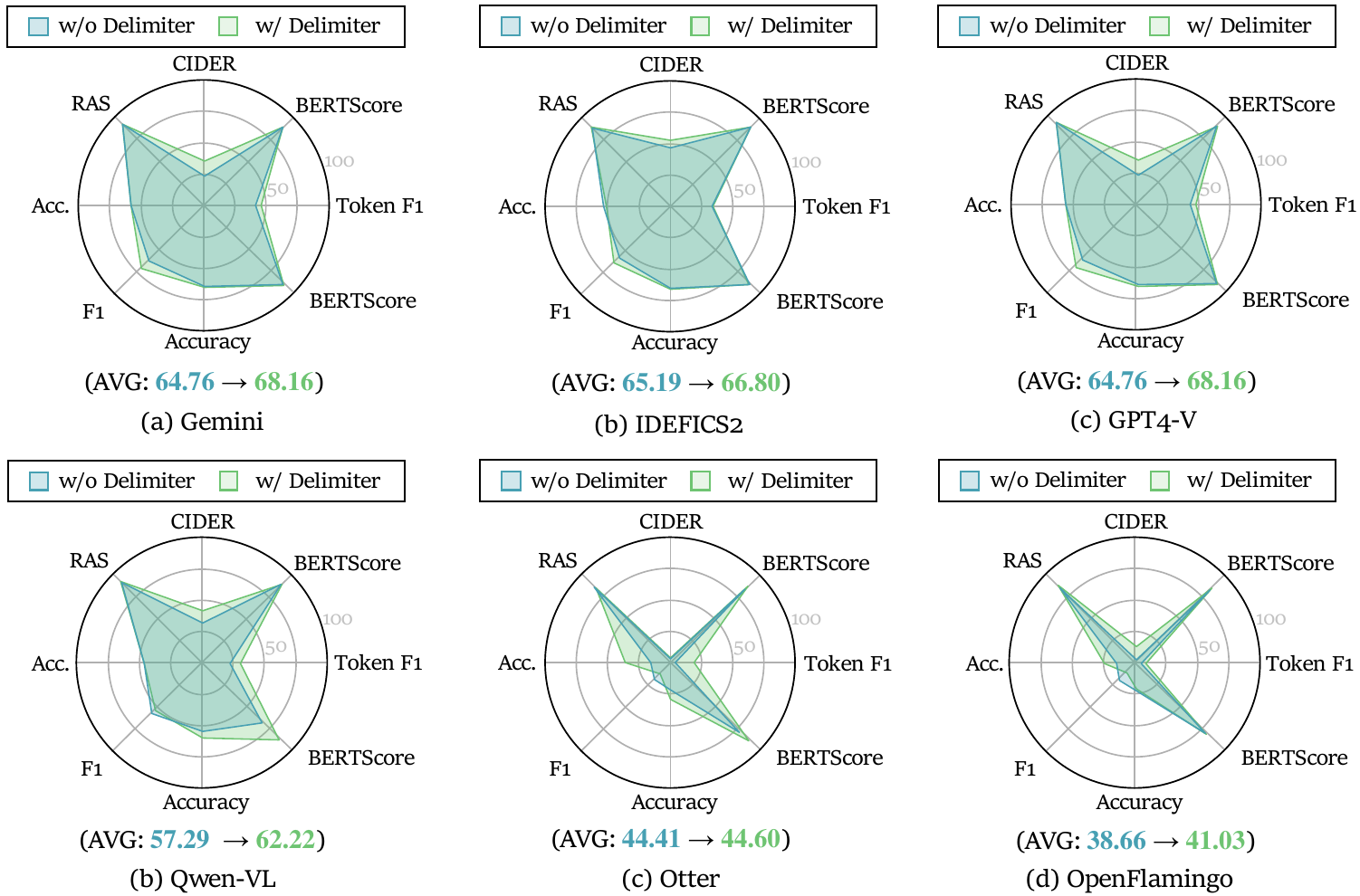}
	\caption{
		The impact of inserting delimiter into the input and output of demonstration on model performance.
	}
	\label{fig:tag-all}
\end{figure}

\subsubsection{The Implement Details for Instruction Injection}
\label{append:instruction}

Visual Language Models (VLLMs) are known to be highly sensitive to input instructions, as demonstrated by \citet{kojima2022large} and \citet{qin2023cross}. Inspired by this observation, we aim to enhance task comprehension in Multi-Modal In-Context Learning (MM-ICL) by incorporating various instructions to explore their influence on performance. Formally, we develop instruction methods, denoted as $\mathcal{I}(\cdot)$, which describe the task and are integrated into the prompt construction process. The prompt $\mathcal{P}$ is constructed as follows:
\begin{equation}
	\mathcal{P} = \mathcal{I}(\delta(x_{\sigma_1}), \delta(x_{\sigma_2}), \ldots, \delta(x_{\sigma_k})),
\end{equation}
where $\delta(x_{\sigma_i})$ represents the transformation of the $i$-th demonstration example.

Specifically, we have designed distinct instructions tailored to different types of tasks, ensuring clarity and appropriateness for each unique context.
For image captioning tasks, the prompt is:
\begin{prompt}
	\ \ 
	Please provide a caption for the image following the structure of the provided example.
\end{prompt}
In this context, the objective is to generate descriptive captions that accurately reflect the content and context of the image.
For Visual Question Answering (VQA) tasks, our prompt is:
\begin{prompt}
	\ \ 
	Examine the image and answer the question by closely following the structure shown in the example provided.
\end{prompt}
The VQA tasks require the model to analyze visual content and respond to specific queries. By following the example, users can produce answers that are precise and directly related to the visual stimuli.
For image classification tasks, the prompt is:
\begin{prompt}
	\ \ 
	Carefully review the image and categorize it based on the options provided in \textbf{[REQUEST]}, following the classification format illustrated in the example.
\end{prompt}
Image classification involves categorizing images into predefined classes based on visual content. The provided example demonstrates the expected classification format. 
For chain-of-thought reasoning tasks, the prompt is:
\begin{prompt}
	\ \ 
	Carefully review the given image and the associated text. Utilize the reasoning format illustrated in the provided examples, breaking down your thought process. Ensure that each reasoning step is explicitly connected to observable details in the image or text, and articulate your conclusion in a clear and logical manner.
\end{prompt}
Chain-of-thought reasoning tasks require a more complex interaction between visual and textual information. The prompt encourages users to break down their reasoning process into clear, logical steps, each supported by specific details from the image or text.

Furthermore, we explore three categories of instructions to enhance the MM-ICL process:
\paragraph{Introductory Instruction ($\mathcal{I}_{intro}$)} This instruction provides an overview of the task before presenting any demonstrations. As depicted in Figure~\ref{fig:prompt-construction-2} (a), the introductory instruction $\mathcal{I}_{intro}$ is positioned at the beginning of the ordered demonstration list $\mathcal{L}$. This setup aims to set the context for the subsequent examples. Specifically, the overall prompt template is as follows:
\begin{prompt}
	\ \ 
	\textit{<Instruction $\mathcal{I}_{intro}$>}

	\textbf{[DEMONSTRATIONS]}
	
	\ \ \textbf{[REQUEST]} \textcolor{deepgray}{\% Shot 1}
	
	\ \ \ \ \textit{<Visual Input $\mathcal{I}_1^{vis}$>} \ \ \textit{<Textual Input $\mathcal{I}_1^{txt}$>}
	
	\ \ \textbf{[RESPONSE]}
	
	\ \ \ \ \textit{<Textual Output $\mathcal{I}_1^{vis}$>}
	
	$\cdots$
	
	\textbf{[QUERY]}
	
	\ \ \textbf{[REQUEST]} \textcolor{deepgray}{\% User Query}
	
	\ \ \ \ \textit{<Visual Input $\mathcal{I}_q^{vis}$>} \ \ \textit{<Textual Input $\mathcal{I}_q^{txt}$>}
	
\end{prompt}

\paragraph{Summative Instruction ($\mathcal{I}_{sum}$)} This instruction offers a summary after the examples, guiding the model to apply the learned concepts to real-world problems. As shown in Figure~\ref{fig:prompt-construction-2} (b), the summative instruction $\mathcal{I}$ is added at the end of the demonstration list $\mathcal{L}$. This helps in reinforcing the learning objectives and expected outcomes. Specifically, the overall prompt template is as follows:
\begin{prompt}
	\ \ 
	\textit{<Instruction $\mathcal{I}_{intro}$>}
	
	\textbf{[DEMONSTRATIONS]}
	
	\ \ \textbf{[REQUEST]} \textcolor{deepgray}{\% Shot 1}
	
	\ \ \ \ \textit{<Visual Input $\mathcal{I}_1^{vis}$>} \ \ \textit{<Textual Input $\mathcal{I}_1^{txt}$>}
	
	\ \ \textbf{[RESPONSE]}
	
	\ \ \ \ \textit{<Textual Output $\mathcal{I}_1^{vis}$>}
	
	$\cdots$
	
	In summary, \textit{<Instruction $\mathcal{I}_{sum}$>}
	
	\textbf{[QUERY]}
	
	\ \ \textbf{[REQUEST]} \textcolor{deepgray}{\% User Query}
	
	\ \ \ \ \textit{<Visual Input $\mathcal{I}_q^{vis}$>} \ \ \textit{<Textual Input $\mathcal{I}_q^{txt}$>}
	
\end{prompt}

\paragraph{Intra-demonstration Instruction ($\mathcal{I}_{intra}$)} This instruction embeds task instructions within each example, assisting the model in understanding the task requirements during the learning process. As illustrated in Figure~\ref{fig:prompt-construction-2} (c), the intra-demonstration instruction $\mathcal{I}$ is included within each demonstration $x_i$ in the list $\mathcal{L}$. This method ensures that the task instructions are continuously reinforced throughout the learning process. Specifically, the overall prompt template is as follows:
\begin{prompt}
	\ \ 
	\textbf{[DEMONSTRATIONS]}
	
	\ \ \textbf{[REQUEST]} \textcolor{deepgray}{\% Shot 1}

	\ \ \ \ \textit{<Visual Input $\mathcal{I}_1^{vis}$>} \ \ \textit{<Textual Input $\mathcal{I}_1^{txt}$>} \textit{<Instruction $\mathcal{I}_{intra}$>}

	\ \ \textbf{[RESPONSE]}
	
	\ \ \ \ \textit{<Textual Output $\mathcal{I}_1^{vis}$>}

	$\cdots$
	
	\textbf{[QUERY]}

	\ \ \textbf{[REQUEST]} \textcolor{deepgray}{\% User Query}
	
	\ \ \ \ \textit{<Visual Input $\mathcal{I}_q^{vis}$>} \ \ \textit{<Textual Input $\mathcal{I}_q^{txt}$>} \textit{<Instruction $\mathcal{I}_{intra}$>}

\end{prompt}

By systematically incorporating these instruction categories into the MM-ICL framework, we aim to investigate their impact on model performance and task comprehension.

\begin{table*}[t]
	\centering
	\begin{adjustbox}{width=0.98\textwidth}
		\begin{tabular}{l|ccc|ccc}
			\toprule
			\multirow{2}{*}{Model} & \multicolumn{3}{c|}{OKVQA~\citep{marino2019ok}} & \multicolumn{3}{c}{VQA-v2~\citep{goyal2017making}} \\
			\cmidrule{2-7}
			& Accuracy & BERTScore & Token F1 & Accuracy & BERTScore & Token F1 \\
			\midrule
			OpenFlamingo~\citep{awadalla2023openflamingo} & 40.28 & 78.10 & 17.45 & 53.33 & 83.34 & 25.67 \\
			GPT4V~\citep{openai2023gpt4} & 54.28 & 85.97 & 25.23 & 69.69 & 84.89 & 29.18 \\
			IDEFICS2~\citep{laurenccon2024matters} & 55.32 & 87.61 & 27.81 & 71.28 & 87.98 & 35.46 \\
			\bottomrule
		\end{tabular}
	\end{adjustbox}
	\caption{
		The correlation analysis of the indicators and reproducted accuracy. The results are obtained by testing on a subset of the test set.
	}
	\label{append:correlation}
	
\end{table*}
\subsection{Prompt Robust}
In our preliminary experiments, we observed that variations in prompts do not significantly alter the overall conclusions. Specifically, we employed multiple prompts—differing in instructions and delimiters—while maintaining equivalent semantic content but varying linguistic expression. As demonstrated in Table~\ref{append:prompt}, the influence of these different prompts on the results is minimal. This suggests that our findings are robust to changes in prompt formulation, thereby supporting the reliability of the experimental outcomes.

\begin{table*}[h]
	\centering
	\begin{adjustbox}{width=0.95\textwidth}
		\begin{tabular}{lccccccccc}
			\toprule
			\multirow{2}{*}{}  & \multicolumn{2}{c}{Caption} & \multicolumn{2}{c}{VQA} & \multicolumn{2}{c}{Classification} & \multicolumn{2}{c}{Reasoning} & \multirow{2}{*}{AVG}
			\\\cmidrule{2-9}
			& CIDER & BERTScore & Token F1 & BERTScore &  Acc & F1 & Acc &  RAS &
			\\
			\midrule
			P1 & 12.03 & 85.85 & 22.53 & 86.67 & 59.93 & 54.62 & 59.52 & 92.04 & 59.15 \\
			P2 & 14.01 & 86.77 & 23.59 & 86.00 & 58.53 & 53.61 & 61.85 & 91.86 & 59.53 \\
			P3 & 13.91 & 86.92 & 24.70 & 87.63 & 59.74 & 52.14 & 61.89 & 93.05 & 60.00 \\
			P4 & 14.44 & 86.48 & 23.14 & 87.77 & 60.23 & 50.48 & 60.54 & 92.27 & 59.42 \\
			\bottomrule
		\end{tabular}
	\end{adjustbox}
	\caption{
		Performance across different prompts (i.e., P1, P2, P3 and P4). 
	}
	\label{append:prompt}
	
\end{table*}

\newpage
\newpage
\section*{NeurIPS Paper Checklist}

\begin{enumerate}
	
	\item {\bf Claims}
	\item[] Question: Do the main claims made in the abstract and introduction accurately reflect the paper's contributions and scope?
	\item[] Answer: \answerYes{} 
	\item[] Justification: We have present our main claims and outline the paper’s contributions and scope in lines 6-13 of the Abstract and 44-54 of the Introduction.
	
	\item {\bf Limitations}
	\item[] Question: Does the paper discuss the limitations of the work performed by the authors?
	\item[] Answer: \answerYes{} 
	\item[] Justification: We have discussed the limitations of our work in Section~\ref{sec:discussion}.
	
	\item {\bf Theory Assumptions and Proofs}
	\item[] Question: For each theoretical result, does the paper provide the full set of assumptions and a complete (and correct) proof?
	\item[] Answer: \answerNA{} 
	\item[] Justification: Our paper does not involve any proofs or assumptions.
	
	\item {\bf Experimental Result Reproducibility}
	\item[] Question: Does the paper fully disclose all the information needed to reproduce the main experimental results of the paper to the extent that it affects the main claims and/or conclusions of the paper (regardless of whether the code and data are provided or not)?
	\item[] Answer: \answerYes{} 
	\item[] Justification: As shown in Section~\ref{sec:analysis}, Section~\ref{sec:setting} and Appendix, we have provided detailed descriptions and analyses of the experimental setups for all our investigations.

	\item {\bf Open access to data and code}
	\item[] Question: Does the paper provide open access to the data and code, with sufficient instructions to faithfully reproduce the main experimental results, as described in supplemental material?
	\item[] Answer: \answerNo{} 
	\item[] Justification: The code for exploratory prompt work generally does not need to be released, and readers can easily use the prompts we report to directly reproduce the results.

	\item {\bf Experimental Setting/Details}
	\item[] Question: Does the paper specify all the training and test details (e.g., data splits, hyperparameters, how they were chosen, type of optimizer, etc.) necessary to understand the results?
	\item[] Answer: \answerYes{} 
	\item[] Justification: As detailed in Section~\ref{sec:setting}, we have thoroughly described the experimental setups for all our explorations.
	
	\item {\bf Experiment Statistical Significance}
	\item[] Question: Does the paper report error bars suitably and correctly defined or other appropriate information about the statistical significance of the experiments?
	\item[] Answer: \answerYes{} 
	\item[] Justification: Error bars are shown in Figure~\ref{fig:token} and Figure~\ref{fig:sample-selection}, with an explanation of the error variables provided in Section~\ref{sec:setting}. However, we do not report error bars for all tasks due to the high costs associated with human annotation and computational resource consumption.
	
	\item {\bf Experiments Compute Resources}
	\item[] Question: For each experiment, does the paper provide sufficient information on the computer resources (type of compute workers, memory, time of execution) needed to reproduce the experiments?
	\item[] Answer: \answerYes{} 
	\item[] Justification: As detailed in Section~\ref{sec:setting}, we outline the specific model compute resources provided.
	
	\item {\bf Code Of Ethics}
	\item[] Question: Does the research conducted in the paper conform, in every respect, with the NeurIPS Code of Ethics \url{https://neurips.cc/public/EthicsGuidelines}?
	\item[] Answer: \answerYes{} 
	\item[] Justification: We are convinced that we comply with NeurIPS Code of Ethics.

	\item {\bf Broader Impacts}
	\item[] Question: Does the paper discuss both potential positive societal impacts and negative societal impacts of the work performed?
	\item[] Answer: \answerYes{} 
	\item[] Justification:  We have addressed the broader impacts of our work in Section~\ref{sec:discussion}. Additionally, as our research is primarily an empirical exploration and poses no additional social risks, we have not included a discussion on potential harmfulness.
	
	\item {\bf Safeguards}
	\item[] Question: Does the paper describe safeguards that have been put in place for responsible release of data or models that have a high risk for misuse (e.g., pretrained language models, image generators, or scraped datasets)?
	\item[] Answer: \answerNA{} 
	\item[] Justification: The paper poses no such risks.
	
	\item {\bf Licenses for existing assets}
	\item[] Question: Are the creators or original owners of assets (e.g., code, data, models), used in the paper, properly credited and are the license and terms of use explicitly mentioned and properly respected?
	\item[] Answer: \answerNA{} 
	\item[] Justification: The paper does not use existing assets.
	
	\item {\bf New Assets}
	\item[] Question: Are new assets introduced in the paper well documented and is the documentation provided alongside the assets?
	\item[] Answer: \answerNA{} 
	\item[] Justification: The paper does not release new assets.
	
	\item {\bf Crowdsourcing and Research with Human Subjects}
	\item[] Question: For crowdsourcing experiments and research with human subjects, does the paper include the full text of instructions given to participants and screenshots, if applicable, as well as details about compensation (if any)? 
	\item[] Answer: \answerNA{} 
	\item[] Justification: The paper does not involve crowdsourcing nor research with human subjects.
	
	\item {\bf Institutional Review Board (IRB) Approvals or Equivalent for Research with Human Subjects}
	\item[] Question: Does the paper describe potential risks incurred by study participants, whether such risks were disclosed to the subjects, and whether Institutional Review Board (IRB) approvals (or an equivalent approval/review based on the requirements of your country or institution) were obtained?
	\item[] Answer: \answerNA{} 
	\item[] Justification: The paper does not involve crowdsourcing nor research with human subjects.
	
\end{enumerate}
\end{document}